\documentclass{IEEEtran}

\usepackage{times}

\usepackage[numbers]{natbib}
\usepackage{multicol}
\usepackage[bookmarks=true]{hyperref}

\usepackage{accents}
\usepackage{amssymb}
\usepackage{graphicx}
\usepackage{setspace}
\usepackage{natbib}
\usepackage{mathrsfs}

\newcommand{\bfzero}{\mathbf{0}}
\newcommand{\atanb}{\mathrm{arctan2}}

\newcommand{\sD}{\mathscr{D}}
\newcommand{\cC}{\mathcal{C}}
\newcommand{\sC}{\bar{\mathcal{C}}}
\newcommand{\cF}{\mathcal{F}}
\newcommand{\cM}{\mathcal{M}}
\newcommand{\cB}{\mathcal{B}}
\newcommand{\cI}{\mathcal{I}}
\newcommand{\cP}{P}
\newcommand{\cO}{O}

\newcommand{\bfa}{\mathbf{a}}

\newcommand{\bfe}{\mathbf{e}}

\newcommand{\bfu}{\mathbf{u}}
\newcommand{\bfv}{\mathbf{v}}
\newcommand{\bfw}{\mathbf{w}}

\newcommand{\ap}{a_\mathrm{plan}}
\newcommand{\zetap}{\zeta_\mathrm{plan}}
\newcommand{\ac}{a_\mathrm{cur}}
\newcommand{\zetac}{\zeta_\mathrm{cur}}
\newcommand{\an}{a_\mathrm{new}}
\newcommand{\zetan}{\zeta_\mathrm{new}}

\newcommand{\bfM}{\mathbf{M}}

\newcommand{\bfQ}{\mathbf{Q}}
\newcommand{\bfR}{\mathbf{R}}

\newcommand{\bfU}{\mathbf{U}}

\title{Affine trajectory correction for\\ nonholonomic mobile robots}

 \author{Quang-Cuong Pham\\
   {\normalsize Nakamura-Takano Laboratory}\\
   {\normalsize Department of Mechano-Informatics}\\
   {\normalsize Graduate School of Information Science and Technology}\\
   {\normalsize University of Tokyo, Japan}\\
   {\normalsize\texttt{cuong.pham@normalesup.org}}}

\begin{document}

\maketitle

\begin{abstract}
  Planning trajectories for nonholonomic systems is difficult and
  computationally expensive. When facing unexpected events, it may
  therefore be preferable to \emph{deform} in some way the initially
  planned trajectory rather than to re-plan entirely a new one. We
  suggest here a method based on affine transformations to make such
  deformations. This method is exact and fast: the deformations and
  the resulting trajectories can be computed algebraically, in one
  step, and without any trajectory re-integration. To demonstrate the
  possibilities offered by this new method, we use it to derive
  position and orientation correction algorithms for the general class
  of planar wheeled robots and for a tridimensional underwater
  vehicle. These algorithms allow in turn achieving more complex
  applications, including obstacle avoidance, feedback control or gap
  filling for sampling-based kinodynamic planners.
\end{abstract}

\begin{IEEEkeywords}
Nonholonomic Motion Planning, Kinematics, Wheeled Robots, Marine
Robotics
\end{IEEEkeywords}

\section{Introduction}
\label{sec:intro}

A bicycle, a car, an aircraft, or a submarine are but a few examples
of nonholonomic systems. Planning trajectories for such systems is
difficult because, by nature, some of their degrees of freedom can
only be controlled in a coupled manner~(see e.g. \cite{Lau98book} and
references therein). As a consequence, when such systems encounter on
their ways an unexpected event (e.g. a random perturbation of the
system state or of the target state, an unforeseen obstacle, etc.), it
may be more efficient to \emph{deform} in some manner the initially
planned trajectory rather than to re-plan entirely a new one
\cite{KhaX97icra,KN03IJRR,LamX04tr,SeiX10wafr}.

Lamiraux and colleagues \cite{LamX04tr} suggested to iteratively
deform the original path by perturbing infinitesimally the control
inputs at each iteration. However, as underlined by Seiler and
colleagues~\cite{SeiX10wafr}, that method requires re-integrating the
whole trajectory at each iteration, which is computationally
expensive. These authors then described a new method based on Lie
group symmetries, which requires re-integrating only \emph{parts} of
the trajectory.

The Lie groups considered in~\cite{CheX08tr,SeiX10wafr} are in fact
Euclidean (or isometry) groups. We propose here to use larger Lie
groups, namely, \emph{affine} groups, which contain the Euclidean
transformations as subgroups. Using affine transformations allows
making more versatile trajectory corrections. In particular, the
corrections are \emph{exact} and can be computed \emph{algebraically},
\emph{in one step}, which makes iterative deformations~\cite{LamX04tr}
or gradient search~\cite{CheX08tr,SeiX10wafr}
unnecessary. Furthermore, there is no need to re-integrate even a part
of the trajectory. Note that, in contrast with previous works where
the studied systems are invariant under Euclidean transformations
\cite{CheX08tr,SeiX10wafr}, here trajectories and control inputs are
not in general affine-invariant. More technical precautions need
therefore to be taken to define and guarantee the feasibility (or
admissibility) of the deformed trajectories under the system
nonholonomic constraints. In particular, the admissibility conditions
are formulated using differential equations with discontinuous
right-hand sides~\cite{Fil88book}.

In section~\ref{sec:general}, we present the general framework of
affine trajectory correction. We then apply this framework to derive
position and orientation correction algorithms for two classical
examples in nonholonomic mobile robotics: the general class of planar
wheeled robots (sections~\ref{sec:wheeled}) and a tridimensional
underwater vehicle (section~\ref{sec:uwv}). In
section~\ref{sec:appli}, we study some more elaborate applications
including trajectory correction for a car towing trailers, obstacle
avoidance, feedback control and gap-filling techniques.  Finally, in
section~\ref{sec:discussion}, we discuss the advantages and drawbacks
of the presented method, its domain of applicability, and possible
future developments.

A preliminary version \cite{Pha11rss} of the present manuscript
describing position and orientation correction algorithms for the
unicycle, the bicycle and an underwater vehicle was accepted for
presentation at the conference \emph{Robotics: Science and Systems}
2011.

%%%%%%%%%%%%%%%%%%%%%%%%%%%%%%%%%%%%%%%%%%%%%%%%%%%%%%%%%%%%
%%%%%%%%%%%%%%%%%%%%%%%%%%%%%%%%%%%%%%%%%%%%%%%%%%%%%%%%%%%%
%%%%%%%%%%%%%%%%%%%%%%%%%%%%%%%%%%%%%%%%%%%%%%%%%%%%%%%%%%%%

\section{General framework}
\label{sec:general}

%%%%%%%%%%%%%%%%%%%%%%%%%%%%%%%%%%%%%%%%%%%%%%%%%%%%%%%%%%%%
\subsection{Affine spaces and affine transformations}
\label{sec:prelim}

An affine space is a set $\mathbb{A}$ together with a group action of
a vector space $\mathbb{W}$. An element $\bfw\in\mathbb{W}$ transforms
a point $\cP\in\mathbb{A}$ into another point $\cP'$ by
$\cP'=\cP+\bfw, $ which can also be noted
$\overrightarrow{\cP\cP'}=\bfw$.

Given a point $\cO\in\mathbb{A}$ (the origin), an affine
transformation $\cF$ of the affine space can be defined by a couple
$(\bfw,\cM)$ where $\bfw\in\mathbb{W}$ and $\cM$ is a non-singular
endomorphism of $\mathbb{W}$ (i.e. a non-singular linear application
$\mathbb{W}\rightarrow\mathbb{W}$). The transformation $\cF$ operates
on $\mathbb{A}$ by
\[
\forall\cP\in\mathbb{A} \quad \cF(\cP)=\cO+
\cM(\overrightarrow{\cO\cP})+\bfw.
\]
Note that, if $\cP_0$ is a fixed-point of $\cF$, then $\cF$ can be
written in the form
\[
\forall\cP\in\mathbb{A} \quad \cF(\cP)=\cP_0+
\cM(\overrightarrow{\cP_0\cP}).
\]

%%%%%%%%%%%%%%%%%%%%%%%%%%%%%%%%%%%%%%%%%%%%%%%%%%%%%%%%%%%%
\subsection{Admissible trajectories and admissible trajectory
  deformations}
\label{sec:intro-adm}

Let us consider a commanded system of dimension $N$. Suppose that $n$
of the system variables form an affine space. As an example, consider
the unicycle model \cite{Lau98book}
\begin{equation}
  \label{eq:unicycle}
  \left\{
    \begin{array}{ccc}
      \dot{x} & = & v\cos(\theta)\\
      \dot{y} & = & v\sin(\theta)\\
      \dot{\theta} & = & \omega
   \end{array} \right.,
\end{equation}
where $(v,\omega)$ are the system control inputs (or commands) and
$(x,y,\theta)$, the system variables. The $(x,y)$ space can be viewed
as an affine space of dimension $n=2$. We call $(x,y)$ the \emph{base
  variables} and the associated affine space, the \emph{base space}.

We say that a full-space trajectory $\sC(t)_{t\in[0,T]}$
($\sC(t)=(x(t),y(t),\theta(t))$ in the above example) is
\emph{admissible} if one can find a set of admissible commands ($v$
and $\omega$ in the example) that generates $\sC$. A base-space
trajectory $\cC$ ($\cC=(x,y)$ in the example) is admissible if there
exists an admissible full-space trajectory whose projection on the
base space coincides with $\cC$.

Let $\cC(t)_{t\in[0,T]}$ be a base-space trajectory and
$\tau\in[0,T]$, a given time instant. We say that a transformation
$\cF$ occurring at $\tau$ \emph{deforms} $\cC(t)_{t\in[0,T]}$ into
$\cC'(t)_{t\in[0,T]}$ if
\[
\begin{array}{rcl}
 \forall t<\tau &~& \cC'(t)=\cC(t)\\
 \forall t\geq\tau &~&\cC'(t)=\cF(\cC(t)).
\end{array}
\]

Given an admissible base-space trajectory $\cC$, an affine
transformation $\cF$ is said to be admissible if $\cF$ deforms $\cC$
into an admissible trajectory.

% Naturally, not every transformation $\cF$ deforms an admissible
% base-space trajectory into another admissible base-space
% trajectory. For instance, in the unicycle model, any admissible
% trajectory $\cC(t)_{t\in[0,T]}$ in the base space corresponds to a
% \emph{continuous} trajectory in the full space. Therefore any admissible
% transformation of $\cC(t)_{t\in[0,T]}$ must deform
% $\cC(t)_{t\in[0,T]}$ into a trajectory whose full-space counterpart is
% continuous, in particular, at the time instant~$\tau$ when the
% deformation occurs.

%%%%%%%%%%%%%%%%%%%%%%%%%%%%%%%%%%%%%%%%%%%%%%%%%%%%%%%%%%%%
\subsection{Differential equations with discontinuous right-hand sides}
\label{sec:disc}

\begin{figure}[ht]
  \centering
  \includegraphics[width=5cm]{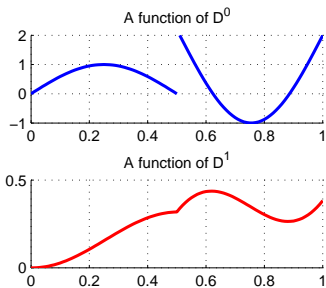}
  \caption{Examples of functions of $\sD^0$ (top) and of $\sD^1$
    (bottom). Note that the top function is actually the derivative of
    the bottom function.}
  \label{fig:ex}
\end{figure}

For convenience, we denote by $\sD^0$ the space of \emph{piecewise
  continuous functions with finite limits at the discontinuity points}
-- or piecewise $C^0$ functions (see Fig.~\ref{fig:ex}, top plot, for
an example). Typically, the linear acceleration of a mobile robot
would belong $\sD^0$: indeed, any brusque press on the throttle or on
the brake pedal would correspond to a discontinuity of the linear
acceleration.

Let $u$ be a function of $\sD^0$ and consider the following
differential equation with discontinuous right-hand side
(see~\cite{Fil88book})
\begin{equation}
  \label{eq:disc}
  \left\{
\begin{array}{rcl}
\dot{x}(t)&=&u(t)\\
x(0)&=&x_0\\
\end{array} \right..
\end{equation}

It follows from the definition of $\sD^0$ that any solution $x$ of
system~(\ref{eq:disc}) is $C^0$ and piecewise $C^1$. Conversely, for
any function $x$ which is $C^0$ and piecewise $C^1$, one has
$\dot{x}\in\sD^0$. For convenience, we denote by $\sD^1$ the space of
such $x$ functions (see Fig.~\ref{fig:ex}, bottom plot, for an
example).

Finally, we denote by $\sD^2$ the space of differentiable functions
whose derivatives are in $\sD^1$. This definition does not involve
technical difficulties since the functions of $\sD^1$ are continuous.

In the unicycle example of section~\ref{sec:intro-adm}, if the linear
and angular accelerations $a$ and $\omega$ are assumed to be in
$\sD^0$ then the linear velocity $v$ and the orientation $\theta$ will
belong to $\sD^1$, which implies in turn that the position $(x,y)$
belongs to $\sD^2$.

%%%%%%%%%%%%%%%%%%%%%%%%%%%%%%%%%%%%%%%%%%%%%%%%%%%%%%%%%%%%
\subsection{Dimension of the space of admissible affine deformations}
\label{sec:affinedef}

From the previous section, one can see that, typically, some of the
variables are required to be \emph{continuous}. These continuity
conditions are particularly critical at the time instant $\tau$ when
the deformation occurs. In general, if one needs to guarantee the
continuities of $m$ variables at $\tau$, this will define $m$
constraints on the set of admissible affine transformations. On the
other hand, the affine transformations of an $n$-dimensional space
form a Lie group of dimension $n+n^2$ ($n$ coordinates for the
translation and $n^2$ coordinates for the endomorphism of the
associated vector space, where $n$ is the number of base
variables). Consequently, if $n+n^2>m$, one could expect to have at
our disposal $\tau$ and $n+n^2-m$ ``extra degrees of freedom'' to
achieve the desired correction while staying admissible.

For wheeled robots of class I (see section~\ref{sec:class1}) and
wheeled robots of class II (section~\ref{sec:class2}), the base space
$(x,y)$ is of dimension $n=2$. We show that there are respectively
$m=4$ and $m=5$ continuity conditions for these systems, yielding
respectively $n+n^2-m=2$ and $n+n^2-m=1$ ``extra degrees of
freedom''. We then suggest how to play with $\tau$ and these ``extra
degrees of freedom'' to make corrections towards virtually any desired
final position and orientation. For the tridimensional underwater
vehicle (section~\ref{sec:uwv}), the base space $(x,y,z)$ is of
dimension $n=3$ and there are $m=6$ continuity conditions, yielding
$n+n^2-m=6$ ``extra degrees of freedom''.

%%%%%%%%%%%%%%%%%%%%%%%%%%%%%%%%%%%%%%%%%%%%%%%%%%%%%%%%%%%%
%%%%%%%%%%%%%%%%%%%%%%%%%%%%%%%%%%%%%%%%%%%%%%%%%%%%%%%%%%%%
%%%%%%%%%%%%%%%%%%%%%%%%%%%%%%%%%%%%%%%%%%%%%%%%%%%%%%%%%%%%

\section{Affine trajectory correction for planar wheeled robots}
\label{sec:wheeled}

The above presented framework suggests the following general scheme to
study affine trajectory correction for a particular system
\begin{enumerate}
\item check the conditions for a base-space trajectory to be
  admissible;
\item characterize the set of admissible affine deformations;
\item compute the admissible affine deformation that achieves the
  desired trajectory correction.
\end{enumerate}

To illustrate, let us now apply the above scheme to wheeled robots,
which constitute an important class of nonholonomic systems.

\subsection{Model description}

At the kinematic level, any wheeled robot whose wheels obey the
rolling without slipping constraints can be modeled by \cite{CamX96tr}
\[
\left\{\begin{array}{ccl}
\dot \xi&=&B(\xi,\beta)\eta\\
\dot \beta&=&\zeta
\end{array}\right.
\]
where $\xi=(x,y,\theta)^\top$ is the \emph{posture} of the robot and
$\beta=(\beta_1\dots\beta_h)^\top$ contains the steering angles of the
\emph{centered orientable conventional wheels} ($h=0$ if there are no
such wheels). As in the unicycle example of
section~\ref{sec:intro-adm}, we choose $x$ and $y$ to be the base
variables. The base space is thus of dimension~2. The non-base
variables are $\theta$ and $\beta_1\dots\beta_h$. 

Throughout this section~\ref{sec:wheeled}, we assume, to avoid
singularities, that the linear velocity $\sqrt{\dot x^2+\dot y^2}$ of
the robot is always strictly positive.

The commands of the system are given by $\eta$, which contains the
linear velocities of well-defined reference points on the frame of the
robot, and $\zeta$, which contains the rates of change of the steering
angles of the centered orientable wheels. We assume that the commands
obey the following conditions
\begin{itemize}
\item the space of admissible commands $\eta$ is $\sD^1$. This is
  consistent with the fact that the linear accelerations $a$ of the
  reference points, which are the derivatives of $\eta$, are in
  $\sD^0$. The possible discontinuities of $a$ would correspond
  to e.g. brusque presses on the throttle or on the brake pedal;
\item the space of admissible commands $\zeta$ is $\sD^0$. The
  possible discontinuities of $\zeta$ would correspond to e.g. hard
  turns of the steering wheel in a car.
\end{itemize}

\subsection{Admissible base-space trajectories}

As shown in \cite{CamX96tr}, any planar wheeled mobile robot can be
described by one out of the five sets of ``forward'' kinematic
equations of Table~\ref{tab:wheeled}, given a suitable choice of a
reference point and of a basis attached to the robot frame.

\begin{table*}[htp]
  \label{tab:wheeled}
  \caption{Forward and reverse kinematic equations for planar wheeled robots}
  \small
  \centering
    \begin{tabular}{|c|c|c|c|c|}
      \hline
      Type&Examples&``Forward'' kinematic equations (cf \cite{CamX96tr})
      &``Reverse'' equations&Admissibility cond.\\
      \hline
      (3,0)&
      \begin{tabular}{c}
        Omni-\\directional\\ robots
      \end{tabular}&
     \begin{tabular}{c}
        $\begin{array}{ccl}
          (\dot x,\dot y)^\top&=&\bfR(\theta)(\eta_1,\eta_2)^\top \\
          \dot \theta&=&\eta_3
        \end{array}$ \\
        where $\bfR(\theta)=\left(\begin{array}{cc}
            \cos\theta&-\sin\theta\\
            \sin\theta&\cos\theta
          \end{array}\right)$
      \end{tabular}
      &
      $\begin{array}{ccl}
        \theta&\in& \sD^2 \textrm{ (arbitrary)}\\
        (\eta_1,\eta_2)^\top&=&\bfR(\theta)^{-1}(\dot x,\dot y)^\top\\
        \eta_3&=&\dot \theta   
      \end{array}$
      &
      \begin{tabular}{c}
        $(x,y) \in \sD^2$\\
      \end{tabular}
      \\
      \hline
      (2,0)&
      \begin{tabular}{c}
        Two-wheel\\ differential\\ drive
      \end{tabular}&
      $\begin{array}{ccl}
        \dot x&=&-\eta_1\sin\theta\\
        \dot y&=&\eta_1\cos\theta\\
        \dot \theta&=&\eta_2
      \end{array}$  &
      $\begin{array}{ccl}
        \theta&=&\atanb(\dot y,\dot x)\\
        \eta_1&=&\sqrt{\dot x^2+\dot y^2}\\
        \eta_2&=&\dot \theta   
      \end{array}$
      &
      \begin{tabular}{c}
        $(x,y) \in \sD^2$\\
        $\atanb(\dot y,\dot x) \in \sD^2$
      \end{tabular}
      \\
      \hline
      (2,1)&
      Unicycle&
      $\begin{array}{ccl}
        \dot x&=&-\eta_1\sin(\theta+\beta)\\
        \dot y&=&\eta_1\cos(\theta+\beta)\\
        \dot \theta&=&\eta_2\\
        \dot\beta&=&\zeta_1
      \end{array}$  &
      $\begin{array}{ccl}
        \theta&\in& \sD^2 \textrm{ (arbitrary)}\\
        \beta&=&\atanb(\dot y,\dot x)-\theta\\
        \eta_1&=&\sqrt{\dot x^2+\dot y^2}\\
        \eta_2&=&\dot \theta\\
        \zeta_1&=&\dot\beta
      \end{array}$
      &
      \begin{tabular}{c}
        $(x,y) \in \sD^2$\\
      \end{tabular}
      \\
      \hline
      (1,1)&
      \begin{tabular}{c}
        Bicycle,\\ kinematic\\ car
      \end{tabular}&
      $\begin{array}{ccl}
        \dot x&=&-\eta_1L\sin\theta\sin\beta\\
        \dot y&=&\eta_1L\cos\theta\sin\beta\\
        \dot \theta&=&\eta_1\cos\beta\\
        \dot\beta&=&\zeta_1
      \end{array}$
      & 
      $\begin{array}{ccl}
        \theta&=&\atanb(\dot y,\dot x)\\
        \beta&=&\atanb(\dot y/(L\cos\theta),\dot\theta)\\
        \eta_1&=&\sqrt{\dot x^2+\dot y^2}/(L\sin\beta)\\
        \zeta_1&=&\dot\beta
      \end{array}$
      &
      \begin{tabular}{c}
        $(x,y) \in \sD^2$\\
        $\atanb(\dot y,\dot x) \in \sD^2$
      \end{tabular}
      \\
      \hline
      (1,2)&
      \begin{tabular}{c}
        Kludge\\robot\\ (cf \cite{CamX96tr})
      \end{tabular}&
     $\begin{array}{ccl}
        \dot x&=&-\eta_1(2L\cos\theta\sin\beta_1\sin\beta_2\\
        &&+L\sin\theta\sin(\beta_1+\beta_2))\\
        \dot y&=&-\eta_1(2L\sin\theta\sin\beta_1\sin\beta_2\\
        &&-L\cos\theta\sin(\beta_1+\beta_2))\\
        \dot \theta&=&\eta_1\sin(\beta_2-\beta_1)\\
        \dot\beta_1&=&\zeta_1\\
        \dot\beta_2&=&\zeta_2
      \end{array}$
      &
      $\begin{array}{ccl}
        \theta&\in& \sD^2 \textrm{ (arbitrary)}\\
        \beta_1&=&\atanb(\dot x\cos\theta+\dot y\sin\theta,\\
        &&2L\dot\theta-\dot x\sin\theta+\dot y\cos\theta)\\
        \beta_1&=&\atanb(\dot x\cos\theta+\dot y\sin\theta,\\
        &&-2L\dot\theta-\dot x\sin\theta+\dot y\cos\theta)\\
        \eta_1&=&\dot\theta/\sin(\beta_2-\beta_1)\\
        \zeta_1&=&\dot\beta_1\\
        \zeta_2&=&\dot\beta_2
  \end{array}$
  &
  \begin{tabular}{c}
    $(x,y) \in \sD^2$\\
  \end{tabular}
  \\
  \hline
\end{tabular}
\end{table*}

For each type of robot, we now characterize the admissible base-space
trajectories given the spaces of admissible commands assumed in the
previous section. The reader is referred to Table~\ref{tab:wheeled}
for the necessary notations and equations.

\subsubsection{Type (3,0)}
\label{sec:30} 

Consider $(\eta_1,\eta_2,\eta_3)\in\sD^1$. The third ``forward''
kinematic equation ($\dot\theta=\eta_3$) implies that
$\theta\in\sD^2$. The first and the second forward equations then
imply that $x\in\sD^2$ and $y\in\sD^2$.

Conversely, consider a base-space trajectory $\cC=(x,y)\in\sD^2$. One
can choose an arbitrary function $\theta\in\sD^2$ and then compute
$(\eta_1,\eta_2,\eta_3)\in\sD^1$ by the ``reverse'' equations.

In summary, a base-space trajectory of a (3,0) wheeled robot is
admissible if and only if it belongs to $\sD^2$.

\subsubsection{Type (2,0)}
\label{sec:20}

Consider $(\eta_1,\eta_2)\in\sD^1$. As previously, the forward
equations imply that $x$ and $y$ belong to $\sD^2$.

Conversely, consider a base-space trajectory $\cC=(x,y)\in\sD^2$. One
can then compute $\theta$ by the first reverse equation
$\theta=\atanb(\dot y,\dot x)$ where
\[
\atanb(b,a)=\left\{
    \begin{array}{cc}
      \pi/2 & \textrm{if $a=0$ and $b\geq0$}\\
      -\pi/2 & \textrm{if $a=0$ and $b<0$}\\
      \arctan(b/a) & \textrm{if $a\neq 0$}
    \end{array} \right..
\]
Remark that the so-calculated $\theta$ belongs to $\sD^1$, but not
necessarily to $\sD^2$. Next, one can compute $\eta_2$ by the third
reverse equation. For $\eta_2$ to be in $\sD^1$, one would need
$\theta\in\sD^2$. As just remarked, the latter condition is \emph{not
  automatically} guaranteed by $\cC=(x,y)\in\sD^2$. On the other hand,
demanding that $\cC=(x,y)\in\sD^3$ would be unduly restrictive. Thus
the condition $\theta\in\sD^2$ must be specified as an independent
supplementary condition.

In summary, a base-space trajectory $\cC$ of a (2,0) robot is
admissible if and only if it belongs to $\sD^2$, \emph{and} if the
function $\theta$ -- as computed from $\cC$ by the first reverse
equation -- also belongs to $\sD^2$.

Note that these admissibility conditions can also be formulated in
terms of continuity constraints on the path
curvature~\cite{BoiX94inria,FS04tr}.

\subsubsection{Type (2,1)}
\label{sec:21}

Consider $(\eta_1,\eta_2)\in\sD^1$ and $\zeta\in\sD^0$. The third and
fourth forward equations imply that $\theta$ and $\beta$ belong
respectively to $\sD^2$ and $\sD^1$. Next, the first and second
forward equations imply that $x$ and $y$ belong to $\sD^2$.

Conversely, consider a base-space trajectory $\cC=(x,y)\in\sD^2$. One
can choose an arbitrary function $\theta\in\sD^2$ and then compute
successively $\beta\in\sD^1$, $(\eta_1,\eta_2)\in\sD^1$, and
$\zeta\in\sD^0$ by the reverse equations.

In summary, as for (3,0) robots, a base-space trajectory of a (2,1)
robot is admissible if and only if it belongs to $\sD^2$.

\subsubsection{Type (1,1)}
\label{sec:11}

As previously, a necessary condition for the admissibility of a
base-space trajectory is that it belongs to $\sD^2$. Conversely,
consider $\cC=(x,y)\in\sD^2$. The first reverse equation allows to
compute $\theta\in\sD^1$. Remark that, as for $(2,0)$ robots, the
so-calculated $\theta$ does not necessarily belong to $\sD^2$. Next,
$\beta$ can be computed from the second reverse equation. Remark that
the derivative of $\theta$ is used in the computation of $\beta$, such
that $\beta$ belongs to $\sD^0$, but not necessarily to
$\sD^1$. However, in order to compute next $\zeta$, one needs
$\beta\in\sD^1$, and consequently $\theta\in\sD^2$.

In summary, as for (2,0) robots, a base-space trajectory $\cC$ of a
(1,1) robot is admissible if and only if it belongs to $\sD^2$,
\emph{and} if the function $\theta$ -- as computed from $\cC$ by the
first reverse equation -- also belongs to $\sD^2$.

\subsubsection{Type (1,2)}
\label{sec:12}

This type of robots can be treated in the same way as (3,0) and (2,1)
robots. A base-space trajectory of a (1,2) robot is admissible if and
only if it belongs to $\sD^2$.

\subsubsection{Summary}
\label{sec:sum}

Following the previous development, one can divide wheeled robots in
two classes. Class I comprises robots of type (3,0), (2,1), and (1,2),
or in other words, those whose \emph{degrees of maneuvrability}
\cite{CamX96tr} equal 3. A base-space trajectory for robots of
this class is admissible if and only if it belongs to $\sD^2$.

Class II comprises robots of type (2,0) and (1,1), or in other words,
those whose \emph{degrees of maneuvrability} equal 2. A base-space
trajectory $\cC=(x,y)$ for robots of this class is admissible if and
only if it belongs to $\sD^2$ \emph{and} if the function
$\theta=\atanb(\dot y,\dot x)$ also belongs to $\sD^2$.

\textbf{Important remark:} From a computational viewpoint, if one
obtains an admissible base-space trajectory $\cC'(t)_{t\in[0,T]}$ (for
instance by deforming a given $\cC(t)_{t\in[0,T]}$), the reverse
equations allow to easily compute the commands that generate that
trajectory by some differentiations and elementary
operations. $\triangle$

\textbf{Relationship with flatness theory:} Our approach here bears
some resemblance with flatness theory \cite{FliX95ijc}. In both cases,
a reduced set of variables is manipulated (here: the
base variables; in flatness theory: the flat outputs) and the state of
the other variables are subsequently recovered from this reduced set
(here: using the reverse equations). There are however two important
differences. First, in our approach, certain non-base variables, in
some systems, are not computed from the base variables but chosen
arbitrarily: e.g. the orientation $\theta$ in wheeled robots of
class~I (see above) or the roll angle $\phi$ in the underwater vehicle
(see section~\ref{sec:uwv-traj}). Second, in some systems, certain
non-base variables are computed from the base variables using
\emph{integration}: e.g. the orientation $\theta_i$ ($i>0$) of the
trailers (see section~\ref{sec:trailers}). In flatness theory,
\emph{all} non-base variables must be computed from the flat outputs,
and they must be done so using only differentiations and algebraic
operations.

Finally, note that it could be interesting to study affine
deformations of the trajectories of the flat outputs. $\triangle$

\subsection{Class I robots}
\label{sec:class1}

We now characterize the affine deformations that preserve the
admissibility of base-space trajectories for robots of class~I. Using
this characterization, we then suggest practical algorithms for
trajectory correction.

\subsubsection{Admissible deformations}
\label{sec:uni-addef}

Consider an admissible base-space trajectory $\cC(t)_{t\in[0,T]}$ and
an affine deformation $\cF$ occurring at time $\tau$ that deforms
$\cC$ into $\cC'$. In what follows, we note $v=\sqrt{\dot x^2+\dot
  y^2}$ (the linear velocity of the robot) and $\theta=\atanb(\dot
y,\dot x)$ (its orientation). Note that, following
section~\ref{sec:sum}, $\cC$ is admissible if and only if
$(x,y)\in\sD^2$, i.e., if and only if $(v,\theta)\in\sD^1$. Note also
that the function $\theta$ here is not the same as the $\theta$ chosen
arbitrarily in Table~\ref{tab:wheeled}. For instance, the unicycle
described by equations~(\ref{eq:unicycle}) is a in fact a (2,1) robot,
with the following correspondance between the variables
\begin{equation}
  \left\{
    \begin{array}{ccc}
    \theta_\mathrm{robot} &=& 0\\
    \beta_\mathrm{robot} &=& \theta_\mathrm{unicycle}-\pi/2\\
    {\eta_1}_\mathrm{robot} &=& v_\mathrm{unicycle}\\
    {\eta_2}_\mathrm{robot} &=& 0\\
    \zeta_\mathrm{robot} &=& \omega_\mathrm{unicycle}
    \end{array}
    \right..
\end{equation}

One has first, by definition,
$\cC'(t)_{t\in(\tau,T]}=\cF(\cC(t)_{t\in(\tau,T]})$. Since $\cF$ is a
smooth application, it is clear that $\cC'(t)_{t\in(\tau,T]}$ -- note
that the interval is open at $\tau$ -- is in $\sD^2$ if and only if
$\cC(t)_{t\in(\tau,T]}$ is in $\sD^2$.

Regarding the time instant $\tau$, the continuities of $x$ and $y$
impose that $\cF(\cC(\tau))=\cC(\tau)$. Thus $\cF$ can be written in
the form 
\begin{equation}
\label{eq:f}
\forall\cP\in\mathbb{A} \quad \cF(\cP)=\cC(\tau)+
\cM(\overrightarrow{\cC(\tau)\cP}).
\end{equation}

One now needs to guarantee the \emph{continuities} of $v$ and $\theta$
at $\tau$, since the two remaining conditions (differentiability and
finite limits for the derivative) do not depend on the behavior of
$\cC'$ at the discrete point $\tau$, and are therefore already
satisfied by virtue of the smoothness of $\cF$.

Consider the velocity \emph{vector}
$\bfv=(\dot{x},\dot{y})^\top$. Remark that the continuity of this
velocity vector is equivalent to those of $v$ and $\theta$. The
continuity of $\bfv$ means that $\bfv(\tau-)$ and $\bfv(\tau+)$ (where
the signs $-$ and $+$ denote respectively the left and right limits)
are well defined, and that $\bfv(\tau-)=\bfv(\tau+)=\bfv(\tau)$.

Similarly, the continuity of $\bfv'$ would mean
$\bfv'(\tau+)=\bfv'(\tau-)=\bfv(\tau)$. On the other hand, one has
$\bfv'(\tau+)=\cM(\bfv(\tau))$. These equalities together imply
$\cM(\bfv(\tau))=\bfv(\tau)$.

Let us now decompose $\cM$ is the basis
$\{\bfu_\parallel,\bfu_\perp\}$ where
$\bfu_\parallel=(\cos(\theta),\sin(\theta))^\top$ is the unit tangent
vector and $\bfu_\perp=(-\sin(\theta),\cos(\theta))^\top$ is the unit
normal vector. The condition $\cM(\bfv(\tau))=\bfv(\tau)$ is
equivalent to
\begin{equation}
  \label{eq:uni-adm}
  \exists \lambda,\mu \in \mathbb{R} \quad \bfM=\left(
    \begin{array}{rcl}
      1&\lambda\\
      0&1+\mu
    \end{array}
  \right),
\end{equation}
where $\bfM$ is the matrix representing $\cM$ in the basis
$\{\bfu_\parallel,\bfu_\perp\}$. 

In summary, the admissible affine transformations at time $\tau$ form
a Lie group of dimension 2, parameterized by $\lambda$ and $\mu$ in
equation~(\ref{eq:uni-adm})

\subsubsection{Trajectory correction}
\label{sec:trajcor}

We consider only the correction of the final position and assume that
$\tau$ is given. It is possible to achieve more complex corrections
(e.g. correcting the final orientation) or to choose ``optimal''
$\tau$s: these developments are left to the reader.

\begin{figure}[ht]
  \centering
  \includegraphics[height=5cm]{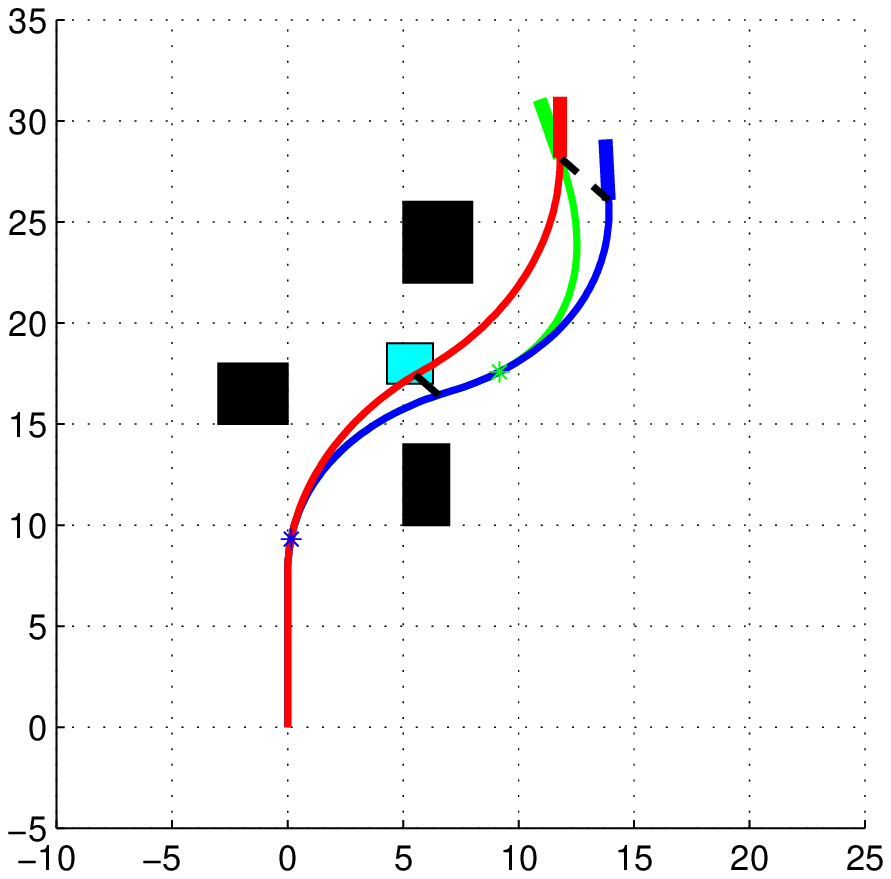}
  \includegraphics[height=5cm]{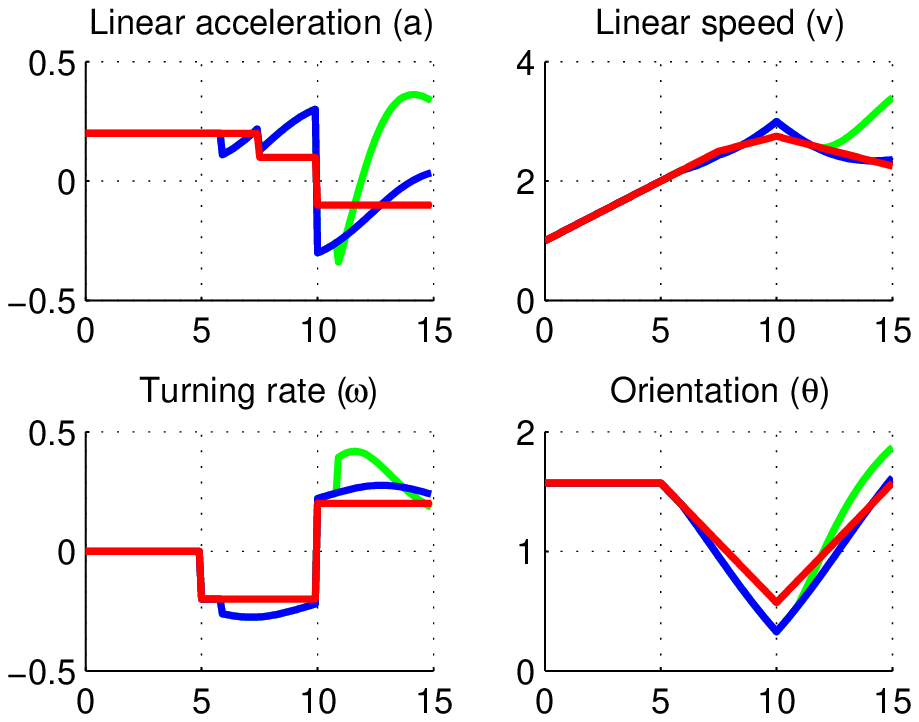}
  \caption{An example of obstacle avoidance for the unicycle. The
    original trajectory $(x,y)$ (red) was planned knowing the position
    of the black obstacles. During the execution, an unforeseen
    obstacle (cyan) appears on the original path. A new trajectory
    $(x_1,y_1)$ (blue) is obtained by deforming the original
    trajectory. The blue star indicates the position
    $(x(\tau_1),y(\tau_1))$ where the deformation occurs, and the
    black plain line joins $(x(t_\mathrm{obs}),y(t_\mathrm{obs}))$ to
    $(x_\mathrm{inter},y_\mathrm{inter})$. Next, in order to get back
    to the original target, an other trajectory (green) is obtained by
    deforming the blue one. The green star indicates the position
    $(x_1(\tau_2),y_1(\tau_2))$ where the deformation occurs, and the
    black dashed line joins $(x_1(T),y_1(T))$ to $(x(T),y(T))$.}
  \label{fig:uni}
\end{figure}

From equation~(\ref{eq:f}), to correct the final position $\cC(T)$
towards a desired position $P_d=(x_d,y_d)$, one needs to look for a
linear application $\cM$ such that
\begin{equation}
\label{eq:m}
\cM(\overrightarrow{\cC(\tau)\cC(T)})=\overrightarrow{\cC(\tau)P_d}.
\end{equation}
Let $\bfQ=[\bfu_\|,\bfu_\perp]$  and let the matrix representing $\cM$
in the basis $\{\bfu_\|,\bfu_\perp\}$ be
\[
\bfM=\left(\begin{array}{cc}
1&\lambda\\
0&1+\mu\\
\end{array} \right).
\]
Equation~(\ref{eq:m}) implies
\begin{equation}
  \label{eq:uni-cond}
  \bfQ\bfM\bfQ^{-1} \left(\begin{array}{c}
x(T)-x(\tau)\\
y(T)-y(\tau)\\
\end{array} \right)
  = \left(\begin{array}{c}
x_d-x(\tau)\\
y_d-y(\tau)\\
\end{array} \right). 
\end{equation}
Let next
\[
\left(\begin{array}{c}
x_1\\
y_1\\
\end{array} \right) =
\bfQ^{-1} \left(\begin{array}{c}
x(T)-x(\tau)\\
y(T)-y(\tau)\\
\end{array} \right),
\]
\[
\left(\begin{array}{c}
x_2\\
y_2\\
\end{array} \right) =
\bfQ^{-1} \left(\begin{array}{c}
x_d-x(\tau)\\
y_d-y(\tau)\\
\end{array} \right). 
\]
Equation~(\ref{eq:uni-cond}) then implies
\[
\lambda=(x_2-x_1)/y_1,
\quad
\mu=(y_2-y_1)/y_1,
\]
provided that $y_1\neq 0$, i.e. that the tangent at $\tau$ does not go
through $\cC(T)$ (see also discussion in
section~\ref{sec:pos}). Fig.~\ref{fig:uni} shows examples of
trajectory corrections for the unicycle.

Note that any desired position in the whole space -- and not only
those in the vicinity of the initially planned final position as in
\cite{SeiX10wafr} -- can theoretically be reached.  Remark on the
other hand that the distance (e.g. the $L_2$ distance) of the
corrected \emph{trajectory} from the original trajectory is a
continuous function of $\lambda$ and $\mu$, meaning that using small
$\lambda$s and $\mu$s results in small changes in the overall
trajectory (and in the commands).

\subsection{Class II robots}
\label{sec:class2}

\subsubsection{Admissible deformations}

Consider an admissible base-space trajectory $\cC$ of a class II robot
and an affine deformation $\cF$ occurring at time $\tau$ that deforms
$\cC$ into $\cC'$. In what follows, we note $v=\sqrt{\dot x^2+\dot
  y^2}$ (the linear velocity of the robot), $\theta=\atanb(\dot y,\dot
x)$ (its orientation), and $\omega=\dot\theta$ (its angular
velocity). Note that, following section~\ref{sec:sum}, $\cC$ is
admissible if and only if $v\in\sD^1$ and $\omega\in\sD^1$.

Following the same reasoning as in section~\ref{sec:uni-addef}, one
can show that $\cC'(t)_{t\in(\tau,T]}$ is in $\sD^2$ if and only if
$\cF(\bfv(\tau))=\bfv(\tau)$, where $\bfv(\tau)$ is is the velocity
vector at $\tau$. One now needs to check the continuities of $\omega'$
at $\tau$ and at the discontinuity points of the second derivative of
$\cC$ (the continuity and differentiability of $\omega'$ elsewhere are
already guaranteed by the smoothness of $\cF$, cf.
section~\ref{sec:uni-addef}).

Consider for this the acceleration vector $\bfa
=(\ddot{x},\ddot{y})^\top$. By definition, one has
\[
\bfa=a\bfu_\parallel+v\omega\bfu_\perp,
\]
with $\bfa$ not necessarily continuous. One can next write
\begin{equation}
  \label{eq:a}
  \bfa\cdot\bfu_\perp=v\omega. 
\end{equation}

Consider now a time instant $t>\tau$ when $\bfa$ is possibly
discontinuous, that is $\bfa(t-)\neq\bfa(t+)$. Since $\omega$ and $v$
are continuous, one has by equation (\ref{eq:a})
\[
\bfa(t-)\cdot\bfu_\perp(t)=\bfa(t+)\cdot\bfu_\perp(t),
\]
or, in other words, that $\bfa(t+)-\bfa(t-)$ and $\bfu_\parallel(t)$
are collinear. Here comes into play a nice property of affine
transformations: they preserve collinearity. Using this property, one
obtains that $\cM(\bfa(t+)-\bfa(t-))$ and $\cM(\bfu_\parallel(t))$ are
collinear. But the former vector is no other than
$\bfa'(t+)-\bfa'(t-)$ and the latter is collinear with
$\bfu'_\parallel(t)$, since
\[
\bfu'_\parallel(t)=\frac{\cM(\bfu_\parallel(t))}{\|\cM(\bfu_\parallel(t))\|}.
\]
Thus $\bfa'(t-)\cdot\bfu'_\perp(t)=\bfa'(t+)\cdot\bfu'_\perp(t)$,
which in turn implies the continuity of $\omega'$ at $t$ (note that this
conclusion also relies on the fact that $v'$ is nonzero if $v$ is
nonzero, owing to the non-singularity of $\cM$).

\textbf{Remark:} Since the affine group is the largest transformation
group of the plane that preserves collinearity, the previous
development shows that it is also the largest group that preserves the
admissibility of every trajectory of a class II robot! $\triangle$

Turning now to the time instant $\tau$, the same
reasoning as previously shows that $\omega'$ is continuous at $\tau$
if and only if
\[
\bfa'(\tau+)\cdot\bfu_\perp(\tau)=\bfa(\tau)\cdot\bfu_\perp(\tau),
\]
or equivalently, if
\begin{equation}
  \label{eq:ii}
  \cM(\bfa(\tau))\cdot\bfu_\perp(\tau)=\bfa(\tau)\cdot\bfu_\perp(\tau).
\end{equation}

Remark now that, since $\bfv\cdot\bfu_\perp=0$, condition
(\ref{eq:ii}) is in fact equivalent to
\[
\exists \lambda\in\mathbb{R} \quad \cM(\bfa(\tau))=\bfa(\tau)+\lambda\bfv(\tau). 
\]

Denoting by $\cB$ the linear application such that
$\cB(\bfv(\tau))=\bfzero$ and $\cB(\bfa(\tau))=\bfv(\tau)$
(one can compute $\cB$ explicitly by
$\cB=[\bfzero,\bfv(\tau)][\bfv(\tau),\bfa(\tau)]^{-1}$), one
obtains
\[
\exists \lambda\in\mathbb{R} \quad \cM=\cI+\lambda\cB,
\]
where $\cI$ is the identity application.

In summary, the admissible affine transformations at time $\tau$ form
a Lie group of dimension 1, given by
$\{\mathcal{I}+\lambda\cB\}_{\lambda\in\mathbb{R}}$.

\textbf{Inflection points:} The previous development is valid only
when $\bfv(\tau)$ and $\bfa(\tau)$ are non-collinear, that is, when
$\cC(\tau)$ is not an \emph{inflection point} (see
also~\cite{BenX09pcb} for an interesting discussion on inflection
points in the context of human movements). $\triangle$

\subsubsection{Trajectory correction I: position correction using one
  affine deformation}
\label{sec:pos}

Let us now play with $\tau$ and the ``extra degree of freedom''
$\lambda$ to make trajectory corrections.

For this, we first study how the final position of the trajectory
$\cC(T)$ is affected by an admissible affine deformation occurring at
time $\tau$. By definition, one has
\[
\begin{array}{rcl}
  \cC'(T)&=&\cC(\tau)+(\cI+\lambda\cB)(\overrightarrow{\cC(\tau)\cC(T)})\\
  &=&\cC(T)+\lambda\cB(\overrightarrow{\cC(\tau)\cC(T)}).
\end{array}
\]

Let us decompose $\overrightarrow{\cC(\tau)\cC(T)}$ in the (in general
non-orthonormal) basis $\{\bfv(\tau),\bfa(\tau)\}$
\[
\overrightarrow{\cC(\tau)\cC(T)}=\gamma \bfv(\tau) + \delta \bfa(\tau).
\]

By definition of $\cB$, one has
\begin{equation}
  \label{eq:endpoint}
  \cC'(T)=\cC(T)+\lambda\delta\bfv(\tau).  
\end{equation}
Consequently, if $\delta$ is nonzero (that is, if
$\overrightarrow{\cC(\tau)\cC(T)}$ and $\bfv(\tau)$ are non-collinear,
or in other words, if the tangent at $\tau$ does not go through
$\cC(T)$), then the locus of $\cC'(T)$ when $\lambda$ varies is the
line that goes through $\cC(T)$ and that collinear with
$\bfv(\tau)$.

In order to make a correction of the final position from $\cC(T)$ to a
desired position $P_d$, it therefore suffices to

\begin{enumerate}
\item compute the vector $\bfe_d=\overrightarrow{\cC(T)P_d}$;
\item find a time instant $\tau$ when the tangent $\bfu_\parallel(\tau)$ is
  collinear with  $\bfe_d$;
\item compute
  $\lambda=\overline{\bfe_d}/(\delta\overline{\bfv(\tau)})$
  where the overline denotes the signed norm;
\item make the affine deformation of parameter $\lambda$ at time
  $\tau$.
\end{enumerate}

\begin{figure}[ht]
  \centering
  \includegraphics[height=5cm]{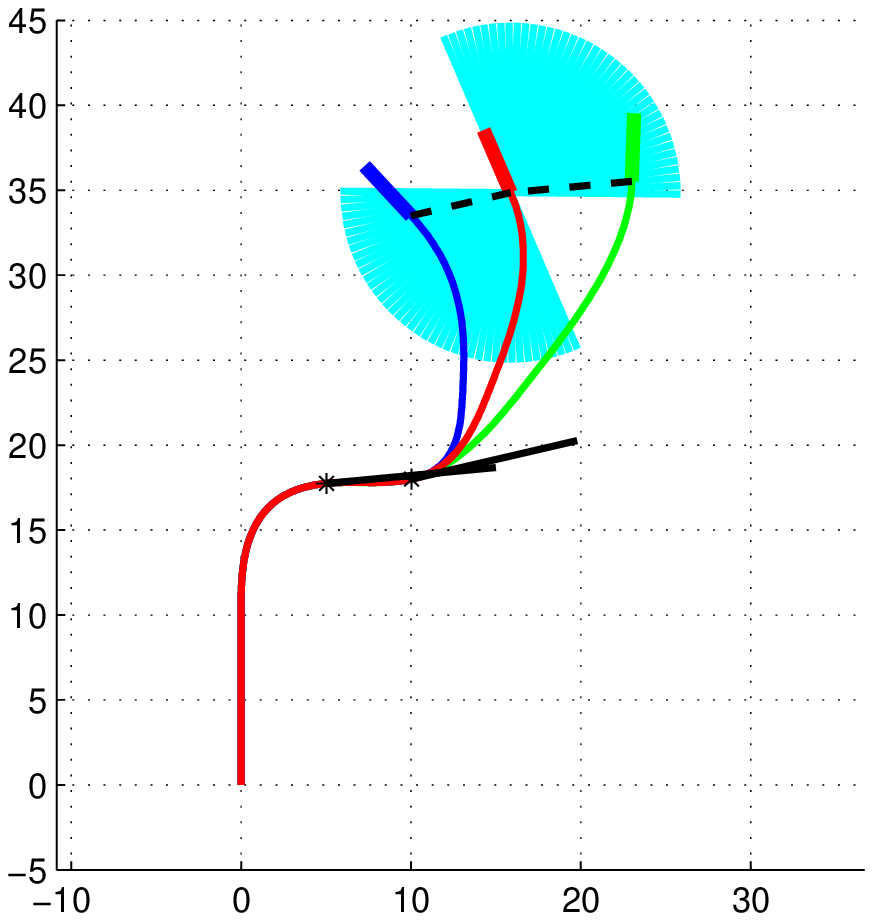}
  \includegraphics[height=5cm]{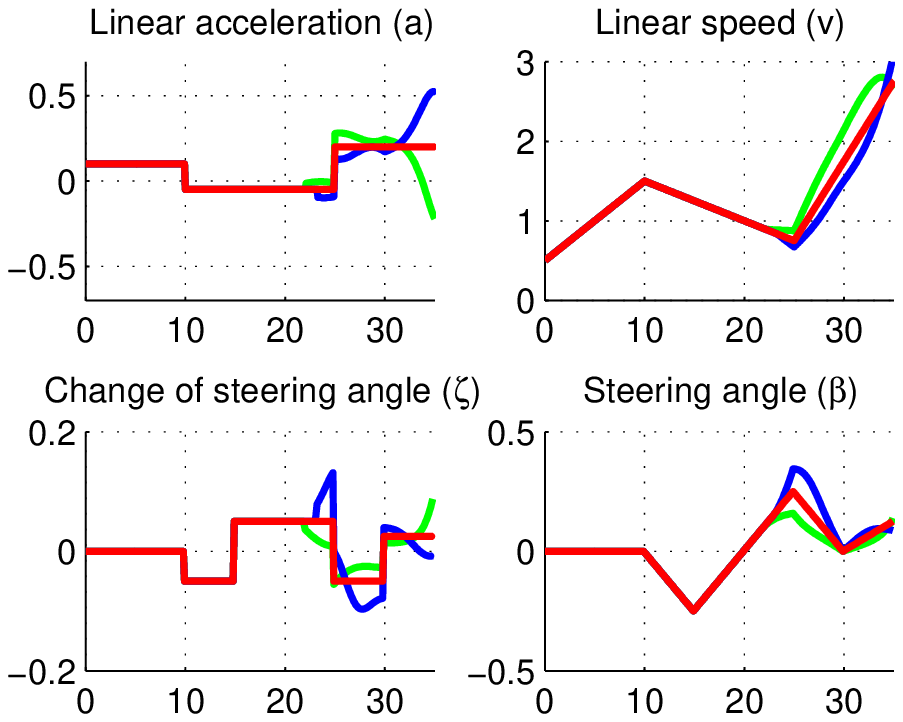}
  \caption{Accessible final positions (in cyan) and two examples of
    position corrections. The original trajectory is in red. For each
    correction, the black plain line represents the tangent at $\tau$
    while the black dotted line joins the original final position
    $\cC(T)$ to the desired final position $P_d$. Note the collinearity
    of the plain line and the dotted line.}
  \label{fig:pos}
\end{figure}

Fig.~\ref{fig:pos} shows some examples of trajectory correction for a
kinematic car, which is a robot of type (1,1). The equation of a
kinematic car is given by~\cite{Lau98book}
\begin{equation}
  \label{eq:kincar}
  \left\{
    \begin{array}{ccc}
      \dot{x} & = & v\cos(\theta)\\
      \dot{y} & = & v\sin(\theta)\\
      \dot{\theta} & = & \frac{v\tan(\beta)}{L}\\
      \dot{\beta}&=&\zeta\\    
    \end{array} \right., 
\end{equation}
which can be put in the form of a robot of type (1,1)
(cf. Table~\ref{tab:wheeled}) using the following correspondance
between the variables
\begin{equation}
  \left\{
    \begin{array}{ccc}
      \theta_\mathrm{robot} &=& \theta_\mathrm{car}-\pi/2\\
      \beta_\mathrm{robot} &=& \pi/2-\beta_\mathrm{car}\\
      {\eta_1}_\mathrm{robot} &=&
      v_\mathrm{car}/(L\cos\beta_\mathrm{car})\\
      \zeta_\mathrm{robot} &=& -\zeta_\mathrm{car}
    \end{array}
  \right..
\end{equation}

% Note that, in a slight contrast with Table~\ref{tab:wheeled}, here the
% steering angle $\beta$ is defined so as to be zero when the car is
% moving on a straight line (while in Table~\ref{tab:wheeled}, one has
% $\beta=\pi/2$ on a straight line).

\textbf{Accessible positions:} From the previous development, it
appears that a position $P_d$ is accessible if and only if the
original trajectory $\cC(t)_{t\in[0,T]}$ has a tangent that is
parallel to $\overrightarrow{\cC(T)P_d}$. Therefore the set of the
trajectory tangents (minus the tangents at the inflection points)
determine the accessible directions for position corrections, as shown
in Fig.~\ref{fig:pos}. $\triangle$

\subsubsection{Trajectory correction II: orientation correction using one
  affine deformation}
\label{sec:or}

Remark that, if $\delta=0$ in equation (\ref{eq:endpoint}), the final
position $\cC(T)$ does not move when $\lambda$ varies. However, the
final \emph{orientation} does vary with $\lambda$. Exploiting this
fact, one can make corrections to the final orientation without
changing the final position.

\begin{figure}[ht]
  \centering
  \begin{minipage}[c]{4cm}
    \centering
    \textbf{A}\\
    \includegraphics[width=4cm]{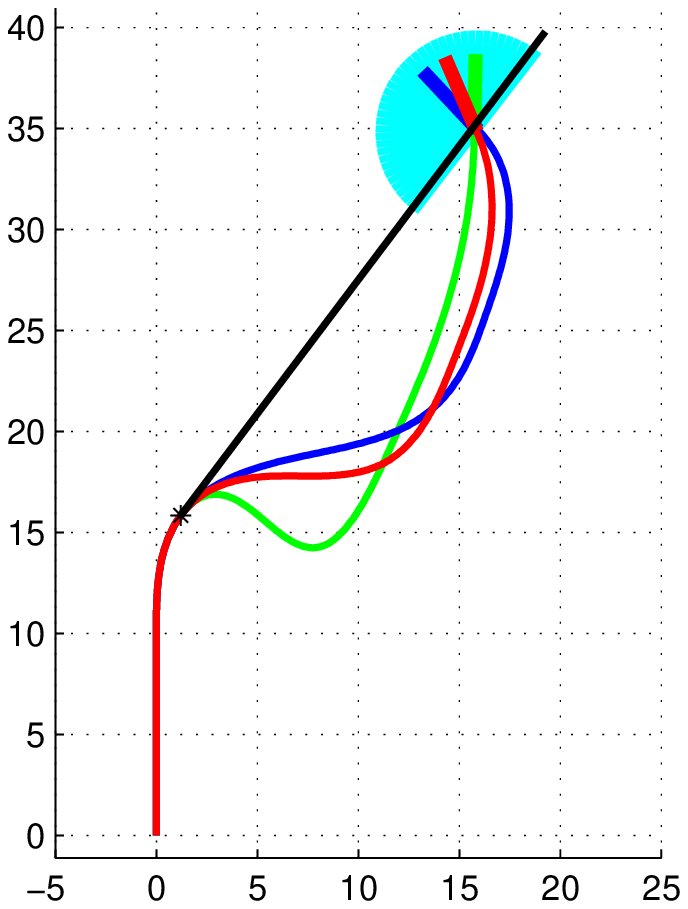}
  \end{minipage}
  \begin{minipage}[c]{3cm}
    \centering
    \textbf{B}\\    
    \includegraphics[width=3cm]{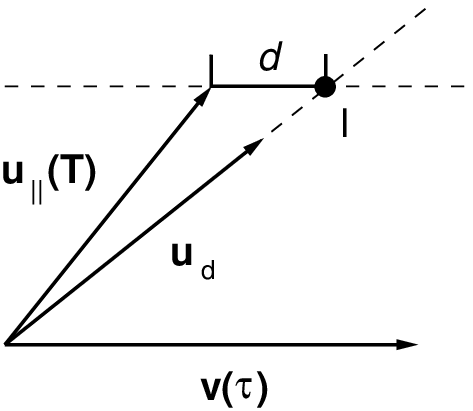}
  \end{minipage}
  \caption{\textbf{A}: accessible final orientations (in cyan) and two
    examples of orientation corrections. The black line represents the
    tangent at $\tau$. Note that the black line goes through the final
    position, which remains unchanged by the orientation
    corrections. \textbf{B}: illustration for the computation of
    $\lambda$ in the correction of the final orientation.}
  \label{fig:or}
\end{figure}

As remarked earlier, $\delta=0$ when $\overrightarrow{\cC(\tau)\cC(T)}$
and $\bfv(\tau)$ are collinear, that is, when the tangent line at time
$\tau$ goes through $\cC(T)$. Consequently, in order to make a
correction of the final tangent vector from $\bfu_\parallel(T)$ to a
desired tangent vector $\bfu_d$ while keeping the final position
unchanged, it suffices to (see Fig.~\ref{fig:or}A)
\begin{enumerate}
\item find a time instant $\tau$ such that the tangent line at $\tau$
  goes through $\cC(T)$;
\item compute the appropriate $\lambda$ (see below);
\item make the affine deformation of parameter $\lambda$ at time
  $\tau$.
\end{enumerate}

\textbf{Computation of $\lambda$:} Remark that the final orientation
of the deformed trajectory is given by the vector
$\cM(\bfu_\parallel(T))$. Observe next that
\[
\cM(\bfu_\parallel(T))=\bfu_\parallel(T)+\lambda\delta_u\bfv(\tau)
\]
where $\delta_u$ is the coefficient multiplying $\bfa(\tau)$ in the
decomposition of $\bfu_\parallel(T)$ in the basis
$\{\bfv(\tau),\bfa(\tau)\}$. 

Consider the intersection $I$ between the line containing $\bfu_d$ and
the line parallel to $\bfv$ and which goes through the tip of
$\bfu_\parallel(T)$ (see illustration in Fig.~\ref{fig:or}B). The
directed distance between $I$ and the tip of $\bfu_\parallel(T)$ is
given by
\[
d=\frac{\sin(\widehat{\bfv(\tau),\bfu_\parallel(T)})}
{\tan(\widehat{\bfv(\tau),\bfu_d})}
-\cos(\widehat{\bfv(\tau),\bfu_\parallel(T)}).
\]
The appropriate $\lambda$ must then satisfy
\[
\lambda\delta_u\overline{\bfv(\tau)}=d,
\]
which leads to $\lambda=d/(\delta_u\overline{\bfv(\tau)})$. $\triangle$

\textbf{Accessible orientations:} The accessible orientations are
restricted to the half-circle defined by the tangent line and in which
lies $\theta(T)$, as shown in Fig.~\ref{fig:or}A. Note
that different choices of the tangent lines (when there exist more
than one possible tangent line) induce different sets of accessible
orientations, whose union forms the total set of accessible
orientations. Note that the tangents at the inflection points are also
forbidden here.~$\triangle$

\subsubsection{Trajectory correction III: position correction using
  \emph{two} affine deformations}

\begin{figure}[ht]
  \centering
  \includegraphics[height=5cm]{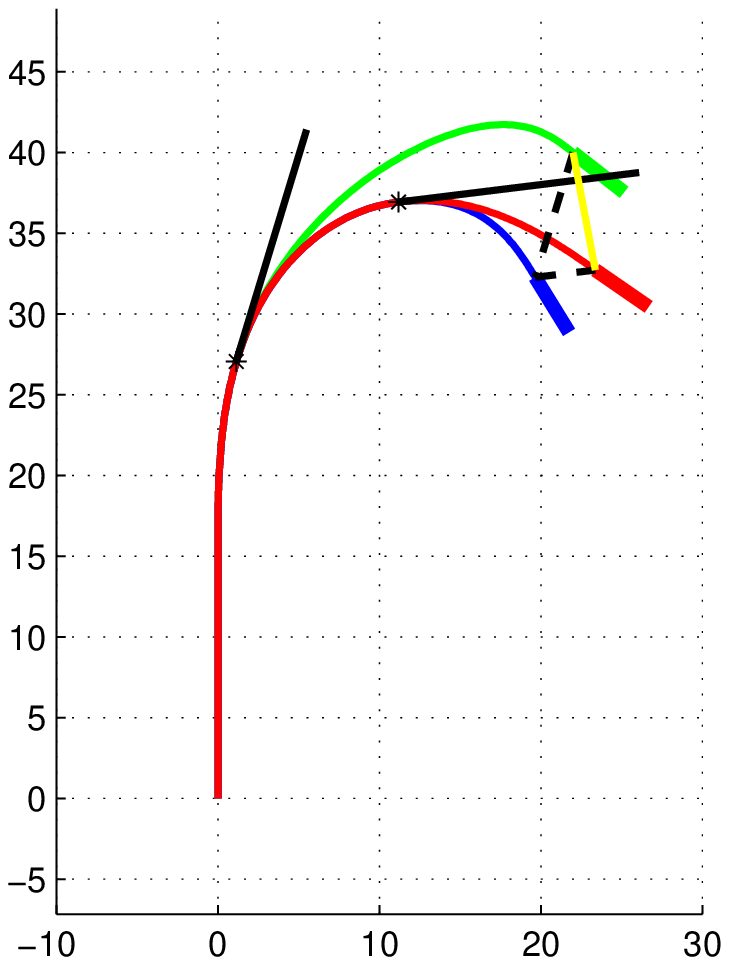}
  \includegraphics[height=5cm]{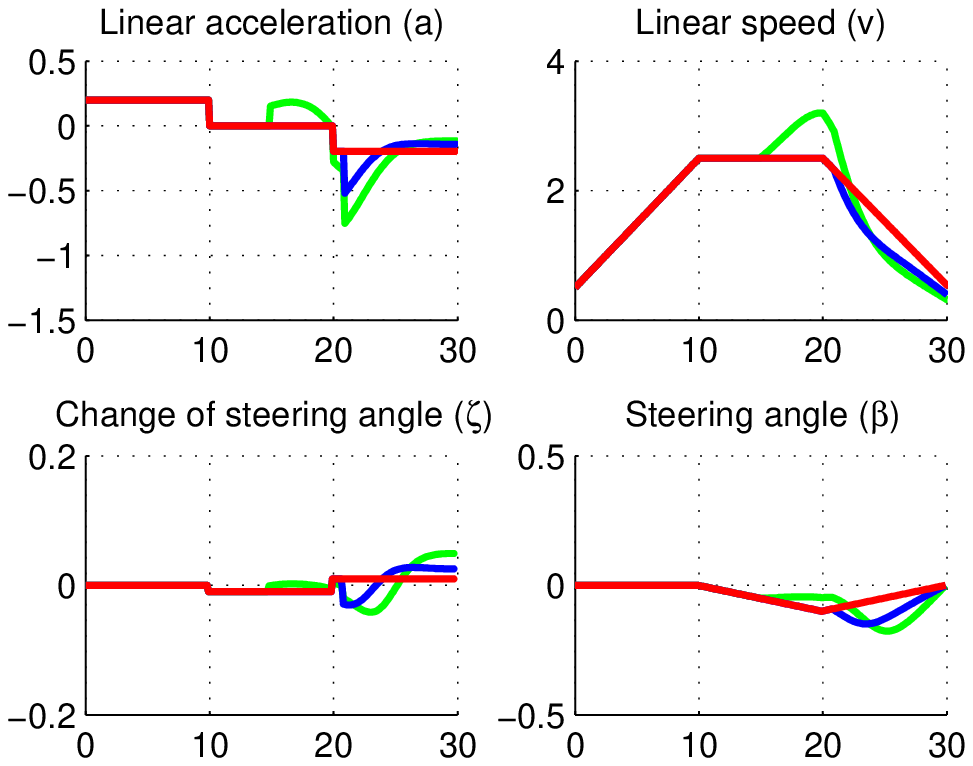}
  \caption{Position correction using two successive affine
    deformations. Note that the desired final position at
    $P_d=(20,40)$ is not accessible by any single deformation
    because the initial trajectory $\cC$ (red) has no tangent parallel
    with the line $\cC(T)P_d$ (yellow line). To overcome this, $\cC$
    is first deformed into $\cC'$ (blue), which in turn is deformed
    into $\cC''$ (green).}
  \label{fig:2steps}
\end{figure}

One can in fact \emph{compose} several affine deformations to achieve
more powerful trajectory corrections. In particular, composing two
deformations allows making position correction towards any desired
final position in space, \emph{so long as the initial trajectory $\cC$
  is not a straight line}, as follows (see Fig.~\ref{fig:2steps})

\begin{enumerate}
\item select two (non-inflection) time instants $\tau_1$ and $\tau_2$,
  with $\tau_1<\tau_2$, such that $\bfv(\tau_1)$ and $\bfv(\tau_2)$
  are non-collinear. Such two time instants exist since $\cC$ is not a
  straight line;
\item decompose $\overrightarrow{\cC(T)P_d}$ in the basis
  $\{\bfv(\tau_1),\bfv(\tau_2)\}$ as $\overrightarrow{\cC(T)P_d}=
  \alpha_1\bfv(\tau_1)+\alpha_2\bfv(\tau_2)$;
\item apply a first deformation on $\cC$ at $\tau_2$ to obtain $\cC'$,
  with $\cC'(T)=\cC(T)+\alpha_2\bfv(\tau_2)$;
\item apply a second deformation on $\cC'$ at $\tau_1$ to obtain
  $\cC''$, with $\cC''(T)=\cC'(T)+\alpha_1\bfv(\tau_1)$. By
  construction
  $\cC''(T)=\cC(T)+\alpha_2\bfv(\tau_2)+\alpha_1\bfv(\tau_1)=P_d$.
\end{enumerate}

It is crucial that the deformation at $\tau_2$ is made first (and
the deformation at $\tau_1$ only second) so as to leave the velocity
vector at $\tau_1$ unchanged ($\bfv'(\tau_1)=\bfv(\tau_1)$).

\subsubsection{Trajectory correction IV: position and orientation
  correction using \emph{three} affine deformations}
\label{sec:3steps}

\begin{figure}[ht]
  \centering
  \includegraphics[height=5cm]{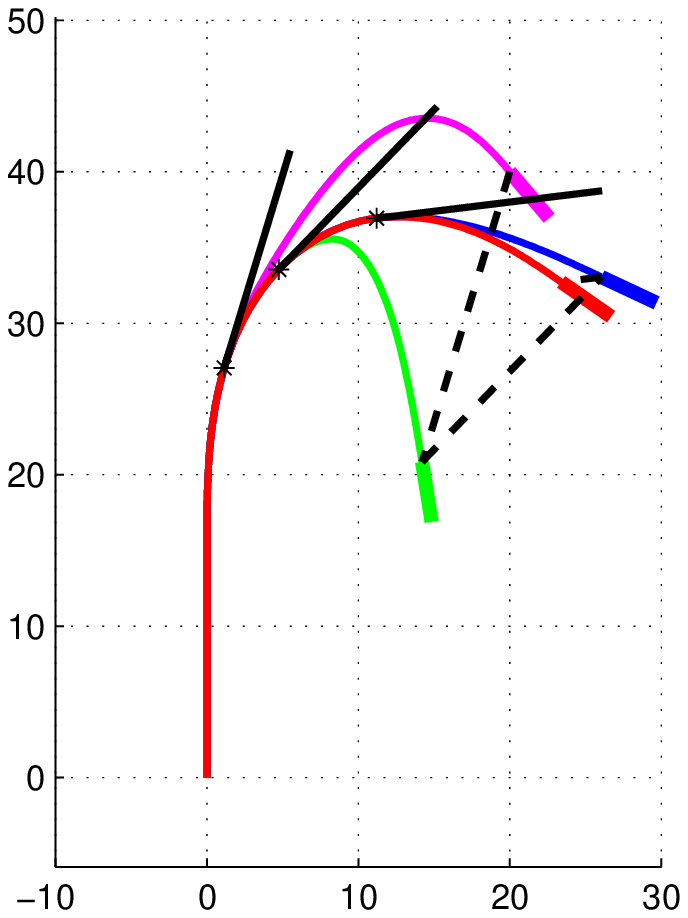}
  \includegraphics[height=5cm]{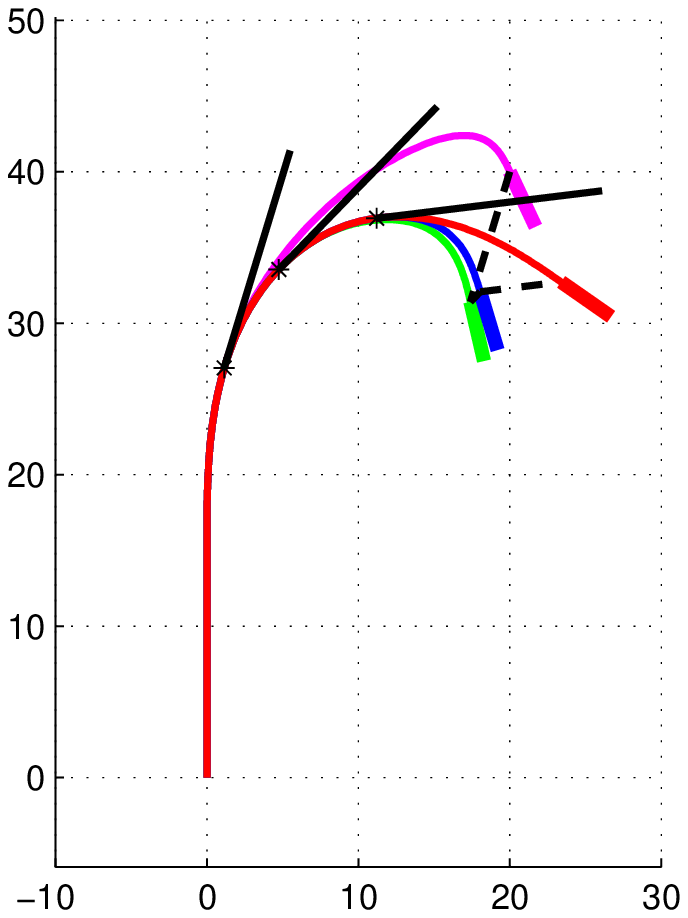}
  \caption{Position and orientation correction using three successive
    affine deformations. The left and right plots correspond to two
    different values of $\alpha_3$. Remark that the trajectory
    $\cC'''$ (magenta) ends at the same position ($P_d=(20,40)$) in
    the left and right plots, but that its final orientation differs
    significantly between the two plots. By varying $\alpha_3$, it is
    thus possible to cover a large range of possible desired final
    orientations while keeping the desired final position fixed.}
  \label{fig:3steps}
\end{figure}

Composing \emph{three} affine deformations allows achieving both
the desired final position and orientation as follows (see
Fig.~\ref{fig:3steps})

\begin{enumerate}
\item select three (non-inflection) time instants $\tau_1$, $\tau_2$,
  and $\tau_3$ with $\tau_1<\tau_2<\tau_3$, such that $\bfv(\tau_1)$,
  $\bfv(\tau_2)$ and $\bfv(\tau_3)$ are pairwise non-collinear. Such
  three time instants exist since $\cC$ is not a straight line;
\item apply a first deformation on $\cC$ at $\tau_3$  to obtain
  $\cC'$, with $\cC'(T)=\cC(T)+\alpha_3\bfv(\tau_3)$, where $\alpha_3$
  is a coefficient to be tuned later;
\item following the results of the previous section, one can use the
  second and third deformations to correct back to the desired
  position $\cC'''(T)=P_d$. Remark that the final \emph{orientation}
  of $\cC'''$ depends on $\alpha_3$ as shown in
  Fig.~\ref{fig:3steps}. The formula to algebraically compute
  $\alpha_3$ as a function of the desired final orientation can be
  obtained in a similar way as in section~\ref{sec:or}.
\end{enumerate}

Finally, remark that one can also set the final linear speed to
arbitrary values while keeping the final position and orientation
unchanged by using the extension technique of section~\ref{sec:gap}.

%%%%%%%%%%%%%%%%%%%%%%%%%%%%%%%%%%%%%%%%%%%%%%%%%%%%%%%%%%%%

\section{Affine trajectory correction for a tridimensional underwater vehicle}
\label{sec:uwv}

\subsection{Model description}

A tridimensional underwater vehicle can be modeled by the following
equations \cite{NS92icra}
\begin{equation}
  \label{eq:uwv}
  \left\{
    \begin{array}{ccc}
      \dot{x} & = & v\cos\psi\cos\theta\\
      \dot{y} & = & v\sin\psi\cos\theta\\
      \dot{z} & = & -v\sin\theta\\
      \left(\begin{array}{c}
          \dot{\phi}\\
          \dot{\theta}\\
          \dot{\psi}
        \end{array} \right)&=&
      \bfR(\phi,\theta) \left(\begin{array}{c}
          \omega_x\\
          \omega_y\\
          \omega_z
        \end{array} \right)
    \end{array} \right.,
\end{equation}
where $(v,\omega_x,\omega_y,\omega_z)$ are the system commands,
$(x,y,z,\phi,\theta,\psi)$ the system variables, and
\[
\bfR(\phi,\theta)=\left(\begin{array}{ccc}
1&\sin\phi\tan\theta&\cos\phi\tan\theta\\
0&\cos\phi&-\sin\phi\\
0&\sin\phi\sec\theta&\cos\phi\sec\theta
\end{array} \right).
\]

We choose $x$, $y$, and $z$ to be the base variables. The base space is
thus of dimension~3. The non-base variables are $\phi$, $\theta$, and
$\psi$.

As in the case of planar wheeled robots, the admissible commands $v$
are assumed to be in $\sD^1$ (allowing possible discontinuities in the
linear acceleration). The admissible commands $\omega_x$, $\omega_y$,
and $\omega_z$ are assumed to be in $\sD^0$.

\subsection{Admissible base-space trajectories}
\label{sec:uwv-traj}

Following the same line of reasoning as previously, a necessary
condition for the admissibility of a base-space trajectory
$\cC(t)_{t\in[0,T]}=(x(t),y(t),z(t))_{t\in[0,T]}$ is that $x$, $y$ and
$z$ belong to $\sD^2$.

Conversely, assume that $x$, $y$ and $z$ belong to $\sD^2$. Remark
first that, from the system equations~(\ref{eq:uwv}), the ``roll''
angle $\phi$ is independent of $(x(t),y(t),z(t))_{t\in[0,T]}$. Next,
given an arbitrary roll angle profile $\phi(t)_{t\in[0,T]}\in\sD^1$,
one can safely write the following reverse equations (assuming that
the velocity is always strictly positive and that the trajectory stays
away from the singularities of the Euler angles \cite{NS92icra})
\[
 \left\{
    \begin{array}{ccc}
      \psi&=&\atanb(\dot{y},\dot{x})\\
      v&=&\sqrt{\dot{x}^2+\dot{y}^2+\dot{z}^2}\\
      \theta&=&\arcsin(\dot{z}/v)\\
      \left(\begin{array}{c}
          \omega_x\\
          \omega_y\\
          \omega_z
        \end{array} \right)&=&
      \bfR(\phi,\theta)^{-1}
      \left(\begin{array}{c}
          \dot{\phi}\\
          \dot{\theta}\\
          \dot{\psi}
        \end{array} \right)
    \end{array} \right.,
\]

In summary, a base-space trajectory is admissible if and only if it is
in~$\sD^2$.

%%%%%%%%%%%%%%%%%%%%%%%%%%%%%%%%%%%%%%%%%%%%%%%%%%%%%%%%%%%%
\subsection{Admissible affine deformations}

Consider now an admissible base-space trajectory $\cC$ and an affine
deformation $\cF$ occurring at time $\tau$ that deforms $\cC$ in to
$\cC'$. As in section~\ref{sec:uni-addef}, one can show that
$\cC'(t)_{t\in(\tau,T]}$ belongs to $\sD^2$, owing to the smoothness
of $\cF$.

At the time instant $\tau$, the continuities of $x$, $y$ and $z$ impose
that $\cF(\cC(\tau))=\cC(\tau)$. Thus $\cF$ can be written in the form
\begin{equation}
  \label{eq:uwv-addef}
  \forall\cP\in\mathbb{A} \quad \cF(\cP)=\cC(\tau)+
  \cM(\overrightarrow{\cC(\tau)\cP}).
\end{equation}

Next, following again the same reasoning as in
section~\ref{sec:uni-addef}, the continuities of $v$, $\psi$ and
$\theta$ are equivalent to setting $\cM(\bfv(\tau))=\bfv(\tau)$.

In summary, an affine deformation $\cF$ occurring at time $\tau$ is
admissible if and only if $\cM(\bfv(\tau))=\bfv(\tau)$ when $\cF$ is
written in the form~(\ref{eq:uwv-addef}).  As a consequence, the
admissible affine transformations at time $\tau$ form a Lie group of
dimension~6.

In practice, we shall compute $\cM$ in the basis
$\{\bfu_\|,\bfw_1,\bfw_2\}$ where $\bfw_1$ and $\bfw_2$ are two
arbitrary unit vectors forming an orthonormal basis with $\bfu_\|$. In
this basis, the condition $\cM(\bfv(\tau))=\bfv(\tau)$ is equivalent
to setting the first column of the matrix that represents $\cM$ to
$(1,0,0)$. It suffices then to find the six remaining coefficients.

\subsection{Trajectory correction}

We consider only the correction of the final position, at a given
$\tau$. It is possible to achieve more complex corrections as well
(correcting the final orientation, avoiding obstacles, etc.) or to
optimize the time instant $\tau$: these developments are left to the
reader.

Theoretically, three free coefficients are sufficient to reach any
final position. As a consequence, we have here more coefficients than
needed. We solve this ``redundancy'' problem by choosing an affine
transformation that is the ``closest'' to the identity matrix, i.e.,
that affects the least the original trajectory.

As in section~\ref{sec:trajcor}, to correct towards a desired position
$P_d=(x_d,y_d,z_d)$, one needs to look for a linear application $\cM$
such that
\begin{equation}
\label{eq:m2}
\cM(\overrightarrow{\cC(\tau)\cC(T)})=\overrightarrow{\cC(\tau)P_d}.
\end{equation}
Let $\bfQ=[\bfu_\|,\bfw_1,\bfw_2]$ and let the matrix representing
$\cM$ in the basis $\{\bfu_\|,\bfw_1,\bfw_2\}$ be
\[
\bfM=\left(\begin{array}{ccc}
1&\lambda&\mu\\
0&1+\nu&\xi\\
0&\sigma&1+\chi
\end{array} \right).
\]
Equation~(\ref{eq:m2}) implies
\begin{equation}
  \label{eq:uwv-cond}
  \bfQ\bfM\bfQ^{-1} \left(\begin{array}{c}
x(T)-x(\tau)\\
y(T)-y(\tau)\\
z(T)-z(\tau)
\end{array} \right)
  = \left(\begin{array}{c}
x_d-x(\tau)\\
y_d-y(\tau)\\
z_d-z(\tau)
\end{array} \right). 
\end{equation}
Let next
\[
\left(\begin{array}{c}
x_1\\
y_1\\
z_1
\end{array} \right) =
\bfQ^{-1} \left(\begin{array}{c}
x(T)-x(\tau)\\
y(T)-y(\tau)\\
z(T)-z(\tau)
\end{array} \right), 
\]
\[
\left(\begin{array}{c}
x_2\\
y_2\\
z_2
\end{array} \right) =
\bfQ^{-1} \left(\begin{array}{c}
x_d-x(\tau)\\
y_d-y(\tau)\\
z_d-z(\tau)
\end{array} \right). 
\]
Equation~(\ref{eq:uwv-cond}) then implies
\begin{equation}
  \label{eq:underdet}
  \bfU
  \left(\begin{array}{c}
      \lambda\\
      \mu\\
      \nu\\
      \xi\\
      \sigma\\
      \chi
    \end{array} \right) =
  \left(\begin{array}{c}
      x_2-x_1\\
      y_2-y_1\\
      z_2-z_1
    \end{array} \right),
\end{equation}
where
\[
\bfU=\left(\begin{array}{cccccc}
      y_1&z_1&0&0&0&0\\
      0&0&y_1&z_1&0&0\\
      0&0&0&0&y_1&z_1
    \end{array} \right).
\]

The $(\lambda,\mu,\nu,\xi,\sigma,\chi)$ with minimal norm (i.e. that
yields a $\cM$ closest to identity according to the Frobenius
distance) and that satisfies the under-determined
system~(\ref{eq:underdet}) is given by
\[
\bfU^+\left(\begin{array}{c}
      x_2-x_1\\
      y_2-y_1\\
      z_2-z_1
    \end{array} \right),
\]
where $\bfU^+$ denotes the Moore-Penrose pseudo-inverse of $\bfU$.

\begin{figure}[ht]
  \centering
  \includegraphics[height=5cm]{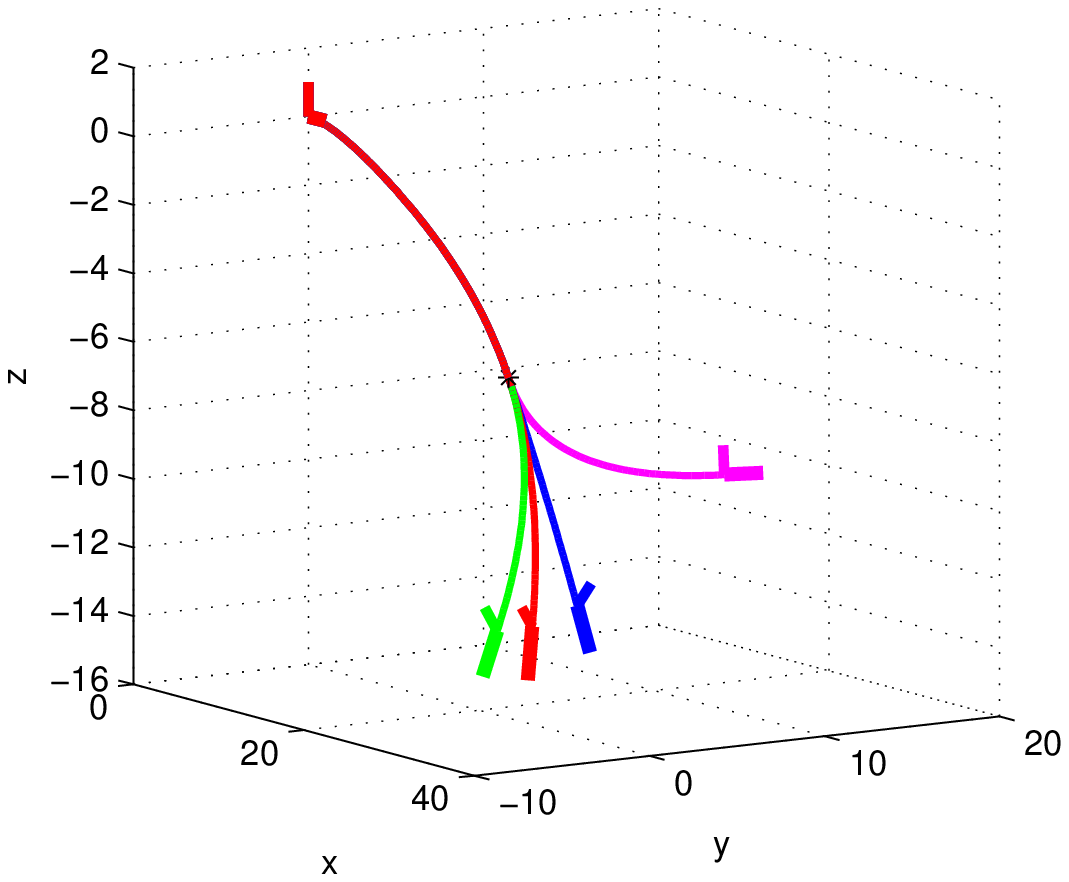}
  \includegraphics[height=6cm]{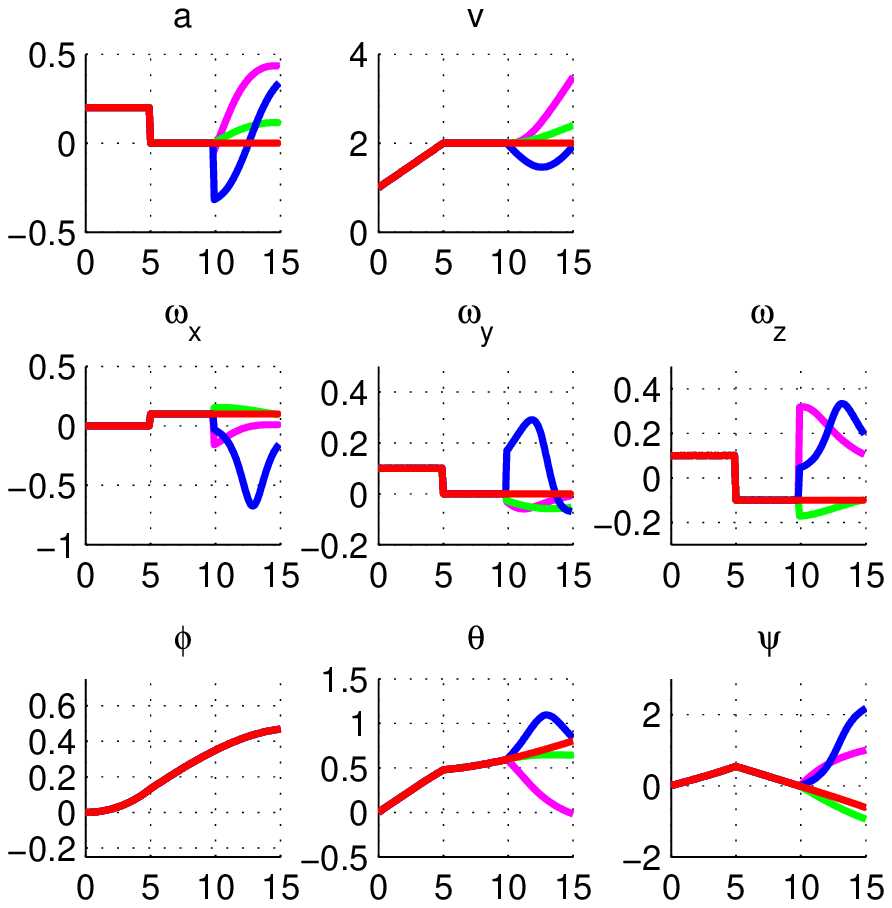}
  \caption{Examples of trajectory corrections for an underwater
    vehicle. The original trajectory is in red.}
  \label{fig:uwv}
\end{figure}

Finally, one needs to choose the ``independent'' angle
$\phi(t)_{t\in[\tau,T]}$. Here our strategy consists of keeping the
same $\phi$ as in the original trajectory. Other strategies (e.g.
keeping the same \emph{absolute} roll as in the original trajectory)
can also be used. Fig.~\ref{fig:uwv} shows some examples of trajectory
corrections.

%%%%%%%%%%%%%%%%%%%%%%%%%%%%%%%%%%%%%%%%%%%%%%%%%%%%%%%%%%%%
%%%%%%%%%%%%%%%%%%%%%%%%%%%%%%%%%%%%%%%%%%%%%%%%%%%%%%%%%%%%
%%%%%%%%%%%%%%%%%%%%%%%%%%%%%%%%%%%%%%%%%%%%%%%%%%%%%%%%%%%%

\section{Further applications}
\label{sec:appli}

We now use the trajectory correction algorithms just developped as
basic tools to tackle more complex tasks. We mostly use the kinematic
car (which is a wheeled robot of type (1,1) and class II, see
section~\ref{sec:11}) as illustrative example but the following
developments can be easily adapted to other nonholonomic systems,
provided that affine corrections are available for these systems.

\subsection{Wheeled robots towing trailers}
\label{sec:trailers}

A kinematic car towing $p$ trailers can be modeled by
\begin{equation}
  \label{eq:trail}
  \left\{
    \begin{array}{ccc}
      \dot{x} & = & v\cos(\theta_0)\\
      \dot{y} & = & v\sin(\theta_0)\\
      \dot{\theta}_0 & = & \frac{v\tan(\beta)}{L_0}\\
      \dot{\beta}&=&\zeta\\
      &&\textrm{and for $i=1\dots p$}\\
      \dot{\theta}_i & = &
      \frac{v}{L_i} \left(\prod_{j=1}^{i-1}{\cos(\theta_{j-1}-\theta_j)}\right)
      \sin(\theta_{i-1}-\theta_i)
    \end{array} \right., 
\end{equation}
where $(v,\zeta)$ are the system commands (respectively the linear
velocity of the car and the rate of change of the steering angle) and
$(x,y,\beta,\theta_0,\theta_1,\dots\theta_n)$, the system variables
(respectively, the $x$ and $y$ coordinates of the car in the
laboratory reference frame, the steering angle, the angle of the car
with respect to the laboratory reference frame, the angle of the first
trailer with respect to the laboratory reference frame, etc.).

\begin{figure}[ht]
  \centering
  \includegraphics[height=5cm]{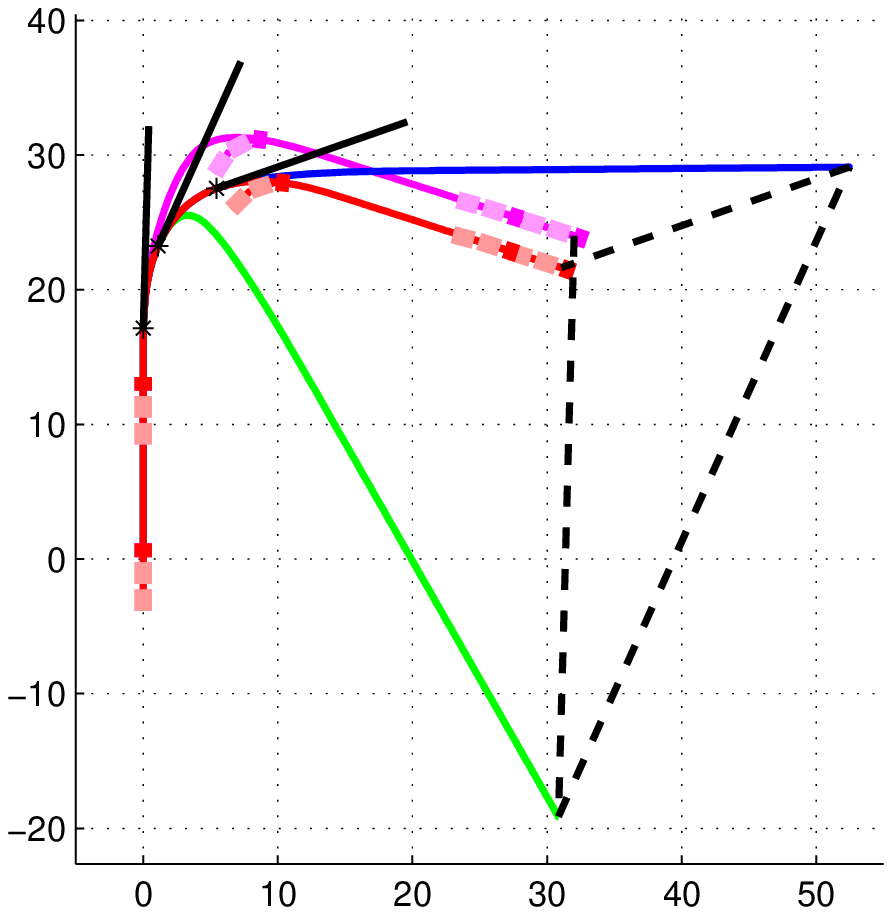}
  \includegraphics[height=6cm]{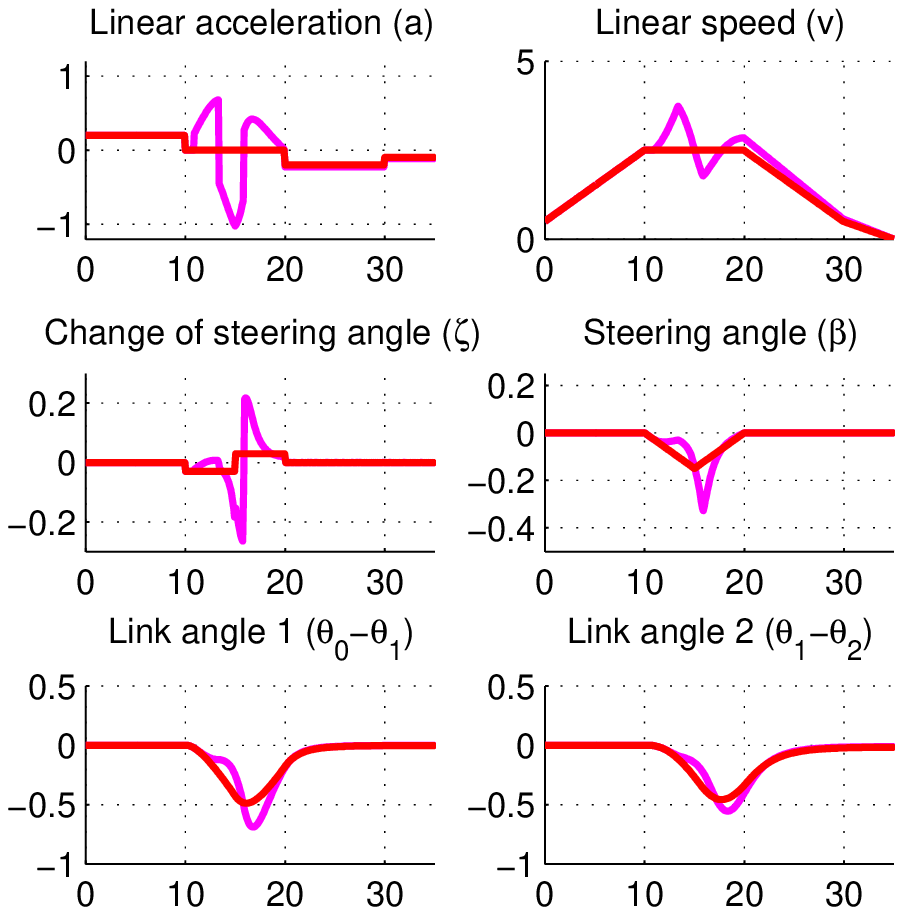}
  \caption{Trajectory correction for a car towing two trailers. One
    can imagine the following scenario: a trajectory $\cC$ (red) is
    initially planned to park the car with trailers in a given parking
    slot; following the occupation of that parking slot, a new
    trajectory $\cC'''$ (magenta) is obtained by deforming the red
    trajectory using the algorithm of section~\ref{sec:3steps} (where
    the blue and green trajectories correspond respectively to $\cC'$
    and $\cC''$). The new trajectory allows the car to be parked in a
    neighboring slot, with the same final orientation. Note that the
    collinearity-preserving property of affine transformations
    automatically guarantees the straightness of the final segments of
    the blue, green and magenta trajectories, which in turn implies
    the alignment of the trailers with the car.}
  \label{fig:trail}
\end{figure}

The same reasoning as in the case of the simple kinematic car shows
that a base-space trajectory $\cC=(x,y)$ is admissible only if it
belongs to $\sD^2$ and if $\theta_0$ -- computed from $\cC$ by
$\theta_0=\atanb(\dot y,\dot x)$ -- belongs to $\sD^2$. Conversely,
assume that $\cC$ is in $\sD^2$ and $\theta_0$ -- computed from $\cC$
-- is in $\sD^2$. One can then safely compute $v\in\sD^1$,
$\beta\in\sD^1$, $\theta_0\in\sD^2$ (by assumption) and
$\zeta\in\sD^0$ as in the case of the simple car. Next, to obtain
$\theta_i$ (for $i=1\dots n$), it suffices to solve successively the
following (ordinary) differential equations
\[
\dot{\theta}_i=\frac{v}{L_i}
\left(\prod_{j=1}^{i-1}{\cos(\theta_{j-1}-\theta_j)}\right)
\sin(\theta_{i-1}-\theta_i)
\]
In summary, the set of admissible base-space trajectories of a car
towing $p$ trailers is the same as that of a simple car. As a
consequence, the admissible affine deformations and the trajectory
correction algorithms for a car towing $p$ trailers are also the same
as those for a simple car.  An example of trajectory correction for a
car towing two trailers is given in Fig.~\ref{fig:trail}.

Note that we have no ``control'' over the configurations of the
trailers, contrary to the literature (transformations to chained forms
\cite{MS93tac,SE95tac}, flatness theory~\cite{FliX95ijc},
etc.). However, consider the (commonly encountered) case when the end
of the initially planned trajectory consists of a straight segment, in
order to align the trailers with the car. Since affine transformations
preserve collinearity, the end of the \emph{corrected} trajectory will
also consist of a straight line, which automatically guarantees the
alignment of the trailers with the car.

Note finally that it could be interesting to study affine deformations
of the trajectory of the flat output~\cite{FliX95ijc}, which is, in
the present case, the middle of the rear wheels axle of the last
trailer (assuming that each trailer is hooked up at the middle of the
rear wheels axle of the preceding trailer).

\subsection{Obstacle avoidance}
\label{sec:obs}

In the trajectory correction algorithms previously developped, one can
in fact replace the final time $T$ by any time instant $t>\tau$. This
allows implementing interactive obstacle avoidance algorithms as
follows

\begin{enumerate}
\item determine a time instant $t_\mathrm{obs}$ when the
  initially planned trajectory would collide with the obstacle;
\item select a new, non colliding, intermediate position
  $(x_\mathrm{inter},y_\mathrm{inter})$ to which one could make
  a correction;
\item make the correction of $(x(t_\mathrm{obs}),y(t_\mathrm{obs}))$
  towards $(x_\mathrm{inter},y_\mathrm{inter})$, using
  $\tau$(s) $<t_\mathrm{obs}$;
\item re-correct the final position towards the initially planned
  final position, using  $\tau$(s) $\geq t_\mathrm{obs}$.
\end{enumerate}

This algorithm can be run iteratively to avoid all obstacles.

One can also prescribe a specific position/orientation of the
trajectory at a given time instant $t_\mathrm{door}$ (this is
desirable for instance when two large obstacles are close to each
other, leaving between them a small doorway through which the robot
could go), as follows

\begin{enumerate}
\item make the correction of $(x(t_\mathrm{door}),y(t_\mathrm{door}))$
  towards the specified intermediate position;
\item make the correction of $\theta(t_\mathrm{door})$ towards the
  specified intermediate orientation;
\item re-correct the final position towards the initially planned
  final position, using  $\tau$(s) $>t_\mathrm{door}$.
\end{enumerate}

\subsection{Feedback control}

So far, we have been focusing on perturbations affecting the state of
the target (position and/or orientation) or the environment
(unexpected appearance of obstacles). Here we show, through a
simplified feedback control algorithm, how affine corrections can also
be used to deal with perturbations affecting the robot's own state.

\begin{figure}[ht]
  \begin{minipage}[c]{4cm}
    \centering
    \textbf{A}\\
    \includegraphics[height=5cm]{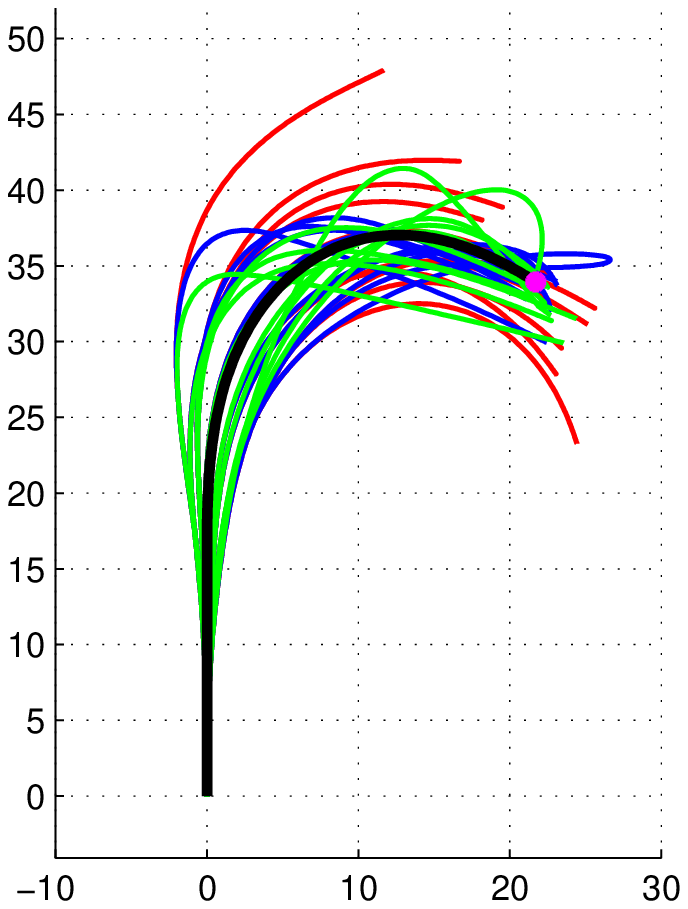}
  \end{minipage}
  \begin{minipage}[c]{3cm} 
    \centering
    \textbf{B}\\    
    \includegraphics[height=3cm]{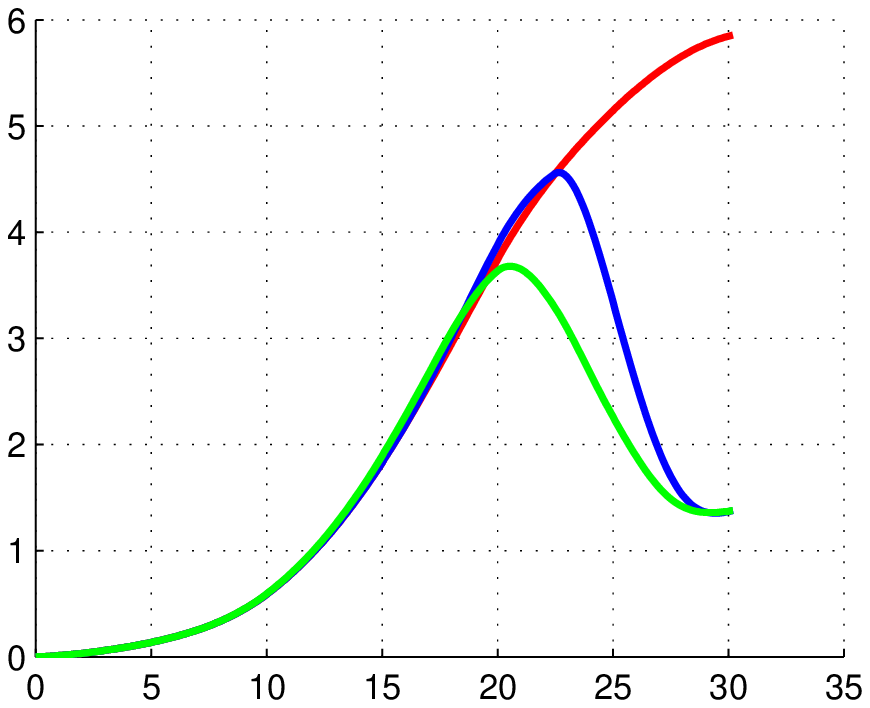}
  \end{minipage}
  \caption{Feedback control using affine corrections. \textbf{A}:
    uncorrected sample trajectories (red), corrected using at most one
    correction (blue) or at most five corrections (green). The
    initially planned trajectory is in black. \textbf{B}: variability
    profiles computed across 2000 realizations of the random
    perturbations $\xi_1$ and $\xi_2$.}
  \label{fig:ofc}
\end{figure}

Consider again the example of the kinematic car. Assume that a
trajectory has been initially planned (black trajectory in
Fig.~\ref{fig:ofc}A), in terms of the time series of the control
inputs $(\ap(t)_{t\in[0,T]},\zetap(t)_{t\in[0,T]})$. Assume now that
these control imputs are \emph{corrupted} by random perturbations
\[
\forall t\in[0,T]\quad 
\left\{
\begin{array}{ccc}
a(t)&=&\ap(t)+\xi_1(t)\\
\zeta(t)&=&\zetap(t)+\xi_2(t)
\end{array}\right.,
\]
where $\xi_1$ and $\xi_2$ two piecewise constant random functions. The
red trajectories in Fig.~\ref{fig:ofc}A represent several trajectories
of the car corresponding to different realizations of the pertubations
$\xi_1$ and $\xi_2$. One can notice that the perturbations make the
final positions of the red trajectories deviate randomly from the
target (denoted by the magenta dot). This can also be noted from the
variability profile (red curve in Fig.~\ref{fig:ofc}B), which is
nonzero at the end of the movement.

We propose the following feedback control algorithm inspired from
\cite{TJ02nat,PH09jnp}. The algorithm maintains at every step two
time series $(\ac(t)_{t\in[0,T]},\zetac(t)_{t\in[0,T]})$ termed
``currently planned control inputs''. These time series are
initialized at the values of
$(\ap(t)_{t\in[0,T]},\zetap(t)_{t\in[0,T]})$.  The movement time $T$
is divided in $S+1$ equal parts. At each time instant $t_i=iT/(S+1)$,
$i=1\dots S$, the robot is given the possibility to make a correction
as follows
\begin{enumerate}
\item compute the final position of the robot, had the control inputs
  $(\ac(t)_{t\in[t_i,T]},\zetac(t)_{t\in[t_i,T]})$ been applied
  starting at the current state $\sC(t_i)$ and until the end of the
  movement. Denote this final simulated position
  $(x_\mathrm{sim},y_\mathrm{sim})$;
\item compute appropriate trajectory deformations with $\tau(s)>t_i$
  to correct the final position from $(x_\mathrm{sim},y_\mathrm{sim})$
  towards $(x_\mathrm{target},y_\mathrm{target})$. This gives rise to
  new time series of control inputs, denoted
  $(\an(t)_{t\in[t_i,T]},\zetan(t)_{t\in[t_i,T]})$;
\item if the new control inputs are acceptable (i.e. do not imply too
  large accelerations or too sharp turns), set
  $\ac(t)_{t\in[t_i,T]}\leftarrow \an(t)_{t\in[t_i,T]}$ and
  $\zetac(t)_{t\in[t_i,T]}\leftarrow
  \zetan(t)_{t\in[t_i,T]}$. Otherwise, keep the current values of
  $\ac$ and $\zetac$.
\end{enumerate}

Figure~\ref{fig:ofc}A shows the results of the feedback control
algorithm for $S=1$ (blue curves) and $S=5$ (green curve). Note that
the blue and green curves are driven by the same realizations of the
perturbations as the red curves (uncorrected trajectories). However,
the blue and green curves end up much closer to the target
position. Figure~\ref{fig:ofc}B confirms this observation: the
final variabilities of the corrected trajectories (blue and green profiles)
at $T$ are much lower ($\sim$1.3m) than that of the uncorrected
trajectories ($\sim$6m).

\begin{figure}[ht]
  \centering
  \includegraphics[height=7cm]{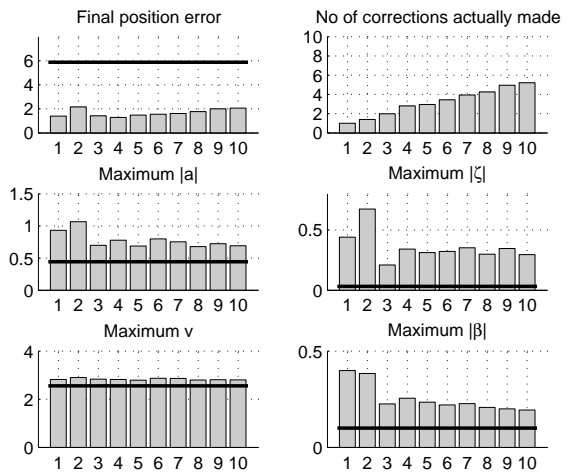}
  \caption{Statistics of the feedback control algorithm across 2000
    realizations of the random perturbations $\xi_1$ and $\xi_2$. The
    X-axis represents the maximum number of corrections allowed
    $S$. The horizontal lines report the values corresponding to the
    uncorrected trajectories ($S=0$).}
  \label{fig:ofc_stats}
\end{figure}

One could ask: why make multiple corrections (green) while making one
unique correction (blue) yields approximately the same final average
error? Figure~\ref{fig:ofc_stats} shows that $S=1$ is associated with
larger values of $a$, $\zeta$ and $\beta$ than $S=5$. This is because
when the robot is allowed to make multiple corrections, the changes to
$a$ and $\zeta$ are \emph{distributed} instead of being concentrated
in one single large correction near the end of the
trajectory. Figure~\ref{fig:ofc}B confirms this observation: the green
variability profile ($S=5$) starts decreasing before the blue
variability profile ($S=1$). Note however that choosing $S>5$ does not
improve the algorithm.

Finally, note that this algorithm is not a trajectory-tracking
algorithm but rather a simplified implementation of ``optimal feedback
control'' \cite{TJ02nat,PH09jnp}.

\subsection{Gap filling for sampling-based kinodynamic planners}
\label{sec:gap}

Gap-reduction techniques are a core component of any sampling-based
kinodynamic planner \cite{CheX08tr}. As an example, consider the
approach proposed in \cite{LK01ijrr}, which consists of
growing two rapidly-exploring random tree (RRT) rooted respectively at
the initial state and at the target state -- a solution trajectory is
obtained when these two trees intersect. When nonholonomic constraints
are present, exact intersections of the trees occur with probability
zero, such that one usually assumes intersection when the trees are
within a nonzero distance of each other, yielding thereby a \emph{gap}
in the solution trajectory. As the performance of the planner
critically depends on the permitted gap size (the larger the permitted
gap size, the quicker the growing trees find an ``intersection'', but
also the more difficult filling the gaps), efficient gap-reduction
techniques have been shown to dramatically improve the performance of
the planner~\cite{CheX08tr}.

\begin{figure}[ht]
  \centering
  \includegraphics[height=5cm]{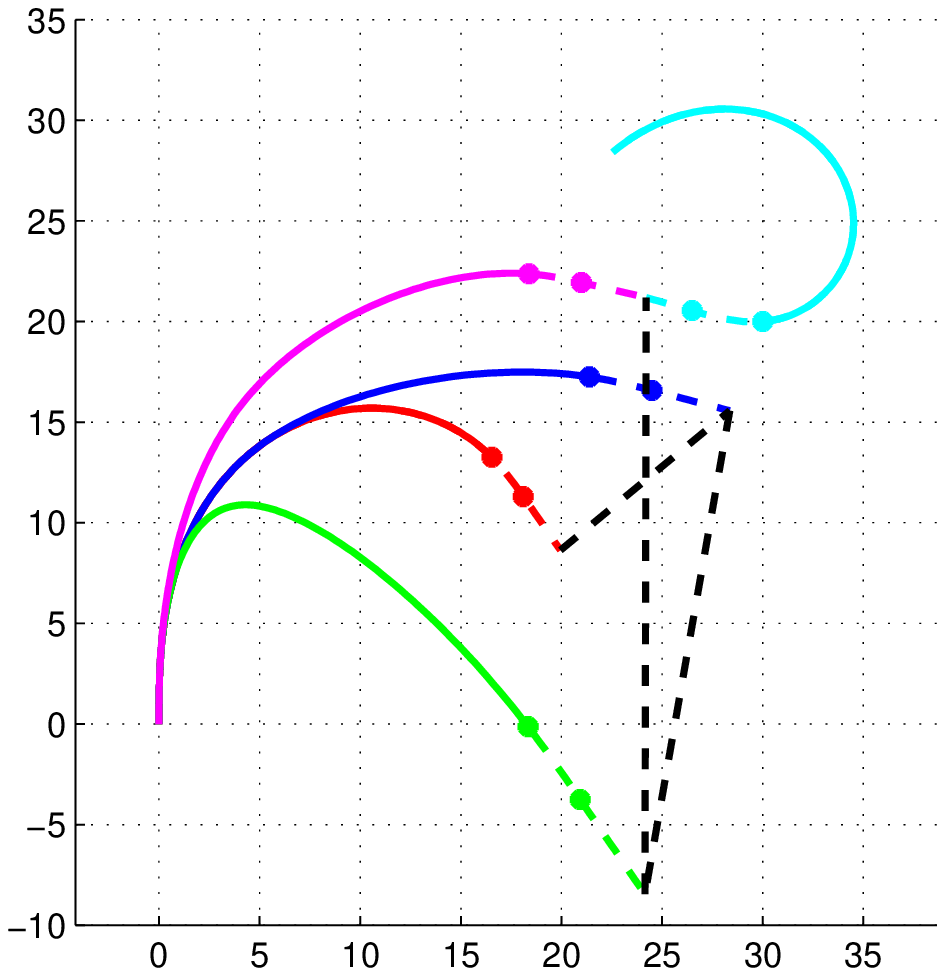}
  \includegraphics[width=8cm]{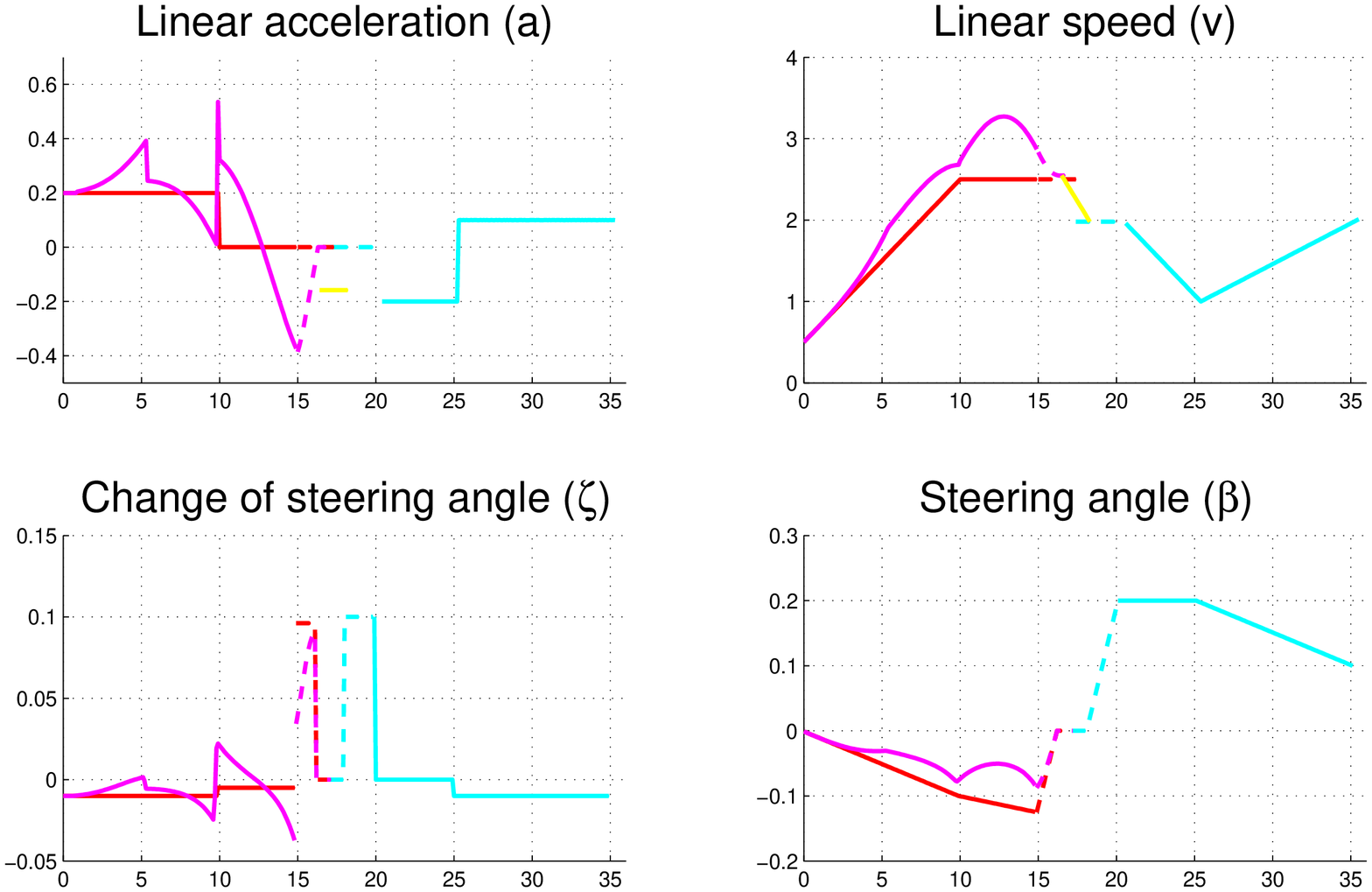}
  \caption{Filling trajectory gaps. Top plot: geometric paths. The
    original trajectories to be connected are showned in plain red
    line ($\cC_1$) and plain cyan line ($\cC_2$). These trajectories
    are first ``prepared'' by growing stubs at their extremities (red
    and cyan dashed lines). The extended $\cC_1$ is then corrected
    into $\cC'''_1$ (magenta) by three successive affine deformations
    (the blue and green lines represent the intermediate trajectories
    $\cC'_1$ and $\cC''_1$ ). Note that $\cC'''_1$ smoothly connects
    with $\cC_2$. Bottom plot: profiles of the other variables. The
    yellow lines in the plots of $a$ and $v$ show the
    modifications that make $v$ continuous without changing the
    geometric paths.}
  \label{fig:gap}
\end{figure}

We now show how affine corrections can be used to fill trajectory
gaps. Consider two trajectories $\cC_1(t)_{t\in[0,T_1]}$ and
$\cC_2(t)_{t\in[0,T_2]}$ of a kinematic car (respectively in red and
cyan in Fig.~\ref{fig:gap}) separated by a gap. We first ``prepare''
the two trajectories as follows
\begin{enumerate}
\item grow a first stub with time duration $\Delta_a$ at the end of
  $\cC_1$. Using the time interval $[T_1,T_1+\Delta_a]$, bring the
  steering angle $\beta_1$ to 0 by ``counter-steering'' (i.e. turning
  the steering wheel back to the straight-ahead position);
\item grow a second stub with time duration $\Delta_b$ at the end of
  the extended $\cC_1(t)$. During this time interval, the steering
  angle $\beta_1$ is kept to 0, resulting in a straight segment. One
  can easily verify that the (doubly) extended trajectory
  $\cC_1(t)_{t\in[0,T_1+\Delta_a+\Delta_b]}$ is admissible. The two
  stubs are shown by dashed red lines in Fig.~\ref{fig:gap};
\item similarly, grow two other stubs at the \emph{beginning} of
  $\cC_2$ (shown by dashed cyan lines in Fig.~\ref{fig:gap}).
\end{enumerate}

After this ``preparation'', we have two trajectories which
respectively ends and begins by straight segments. The lengths of the
added stubs depend on the $\Delta$s and can be made relatively short
if the $\beta$s are small and large braking and counter-steering rates
are permitted. We can now use the position and orientation algorithms
given in the previous sections to bring the end of the extended
$\cC_1$ towards the beginning of the extended $\cC_2$.
Fig.~\ref{fig:gap} shows an example of such correction using three
succesive affine deformations (cf. section~\ref{sec:3steps}). The
admissibility conditions are verified by observing that
\begin{itemize}
\item since affine transformations preserve collinearity, the
  corrected extended trajectory $\cC'''_1$ (magenta) also ends by a
  straight segment. When this straight segment connects with the
  straight segment at the beginning of the extended $\cC_2$, the
  continuity of $\beta$ is guaranteed;
\item regarding the continuity of $v$, one can use the straight parts
  around the connection point to modulate the speed profile to make it
  continuous \emph{without altering the geometric path}: see the
  yellow lines in the plots of $a$ and $v$ in Fig.~\ref{fig:gap}.
\end{itemize}

%%%%%%%%%%%%%%%%%%%%%%%%%%%%%%%%%%%%%%%%%%%%%%%%%%%%%%%%%%%%
%%%%%%%%%%%%%%%%%%%%%%%%%%%%%%%%%%%%%%%%%%%%%%%%%%%%%%%%%%%%
%%%%%%%%%%%%%%%%%%%%%%%%%%%%%%%%%%%%%%%%%%%%%%%%%%%%%%%%%%%%

\section{Discussion}
\label{sec:discussion}

As stated at the beginning of section~\ref{sec:wheeled}, one can
apply the following general scheme to study affine trajectory
corrections for nonholonomic systems
\begin{enumerate}
\item check the conditions for a base-space trajectory to be
  admissible. Often (but not always), a base-space trajectory is
  admissible if it -- and some functions computed from it -- belong to
  certain classes $\sD^i$;
\item based on the admissibility conditions of trajectories,
  particularly at the time instant when the deformation occurs,
  characterize the set of admissible affine deformations. Often (but
  not always), the admissible affine deformations at a given time
  instant form a Lie group of dimension $n+n^2-m$ where $n$ is the
  number of base variables and $m$ the number of continuity
  conditions;
\item finally, play with $\tau$ and the $n+n^2-m$ ``extra degrees of
  freedom'' to achieve the desired correction. If there are more
  ``extra degrees of freedom'' than needed, one can ``optimize'' by
  choosing the affine transformations that are the closest to
  identity.
\end{enumerate}

This general scheme suggests in turn the classes of systems that can
or cannot be tackled by the proposed method. For instance, an
underwater vehicle whose changes in turning rate
($\rho_x=\dot{\omega}_x$, $\rho_y=\dot{\omega}_y$,
$\rho_z=\dot{\omega}_z$) are required to be continuous could probably
be treated by the method (since in this case $n+n^2-m=3$). The
development of the theory to deal with other classes of nonholonomic
systems are also the subject of ongoing efforts.

Holonomic systems, such as the end-point of a robotic manipulator, are
not subject, by nature, to the differential constraints with which the
current manuscript is concerned. However, it is sometimes desirable
for efficiency reasons to \emph{artificially enforce} some
differential constraints, such as the continuity of the velocity
vector. For instance, if a planned \emph{path} is not $\mathscr{C}^1$
at some points, the robot must stop-and-start at these points
\cite{KanX08iros}, which clearly is an undesired behavior. In this
perspective, the regularity-preserving deformation algorithms
developped here can also be useful for holonomic trajectory planning.

As just remarked, this manuscript is mostly concerned with the
differential constraints that stem from the nonholonomic nature of the
considered systems. In practice, other constraints, such as upper
limits on the absolute acceleration or on the trajectory curvature,
could further restrict the set of admissible affine deformations. This
can be treated by observing that the changes in acceleration or
curvature from the original trajectory can be computed from the affine
transformation at hand (see also \cite{BenX09pcb}). The integration of
such constraints into the current framework represents an
important task (see e.g. \cite{HL07icra}).

Another promising direction of research may consist of
\emph{combining} the approach presented here with existing approaches
for trajectory planning and deformation. We have mentioned earlier
possible interactions with flatness theory. A complementary use of
affine-based and perturbation-based deformations \cite{LamX04tr} may
also lead to more efficient algorithms. For instance, affine
corrections perform badly when the original trajectory is close to a
straight line. Using the results in~\cite{LamX04tr}, it should be
possible to slightly perturb the original trajectory to generate local
curved portions, which subsequently allow applying affine deformations
with greater effectiveness.

As mentioned in the Introduction, one advantage of the method
presented in this manuscript is that it requires no re-integration of
the trajectory. On the other hand, \emph{differentiations} of the
trajectory must be performed in order to recover the commands (see
``Important remark'' in section~\ref{sec:sum}). Note however that, if
multiple deformations are made, the differentiations need to be
performed only \emph{once}, after all the deformations have been
applied.

The group property of affine transformations can also be used to
further accelerate the computations (as in \cite{SeiX10wafr} with
Euclidean transformations). Assume for instance that two affine
transformations $\cF_1$ and $\cF_2$ are applied at time instants
$\tau_1$ and $\tau_2$, with $\tau_1<\tau_2$. Then one can apply
$\cF_1$ to $\cC(t)_{t\in[\tau_1,\tau_2]}$ and next $\cF_2\circ\cF_1$,
which is also an affine transformation, to $\cC(t)_{t\in(\tau_2,T]}$.

Another advantage, also mentioned in the Introduction, is that the
method presented here can be executed \emph{in one step}, while other
methods require iterative deformations of the
trajectory~\cite{LamX04tr} or gradient descent to find the appropriate
deformation coefficients~\cite{CheX08tr,SeiX10wafr}. This may result
in significant performance gains, in particular, in real-time
applications or in highly compute-intensive tasks such as the building
of probabilistic roadmaps~\cite{LK01ijrr}.

Finally, the method is exact: for example, a desired position can be
reached \emph{exactly}, and not only approached iteratively ``as close
as we want''. This may have important consequences. For example, in
the \emph{initial} trajectory planning, one would no longer need to
spend time finding a trajectory that ends very close to the
target. Instead, one can plan a trajectory that ends roughly somewhere
near the target, and then make an affine deformation towards the exact
target position.

A last word on the biological implications of the ideas presented
here. One source of inspiration for the present work was indeed the
recent studies of affine invariance in human perception and
movements~(see e.g. \cite{BenX09pcb} and references
therein). Conversely, one could ask (and experimentally test) whether
humans use algorithms similar to those described here to correct their
hand or locomotor trajectories.

\subsection*{Acknowledgments}
The author is deeply grateful to Prof. Daniel Bennequin,
Prof. Yoshihiko Nakamura, and Dr. Oussama Kanoun for their highly
valuable suggestions and comments. This research was funded by an
University of Tokyo grant and by a JSPS postdoctoral fellowship.

\bibliographystyle{plainnat}
\bibliography{../ynl}

\end{document}